\documentclass{article}

\usepackage[preprint]{neurips_2026}

\usepackage[utf8]{inputenc} 
\usepackage[T1]{fontenc}    
\usepackage{hyperref}       
\usepackage{url}            
\usepackage{booktabs}       
\usepackage{amsfonts}       
\usepackage{nicefrac}       
\usepackage{microtype}      
\usepackage{xcolor}         
\usepackage{graphicx}

\usepackage{xcolor}
\usepackage{multirow}
\usepackage{colortbl}
\usepackage[normalem]{ulem}
\useunder{\uline}{\ul}{}
\usepackage[dvipsnames]{xcolor}
\newcommand{\gain}[1]{\textcolor{green!60!black}{\scriptsize #1}}
\usepackage{wrapfig}
\usepackage{amsthm}

\usepackage{amsmath}
\usepackage{amssymb}
\usepackage{mathtools}
\usepackage{enumitem}

\usepackage{algorithm}
\usepackage{float}
\usepackage{subcaption}
\usepackage[noend]{algpseudocode}
\theoremstyle{plain}

\theoremstyle{definition}

\theoremstyle{remark}

\title{REMAP: \underline{Re}gularized \underline{Ma}tching and \underline{P}artial Alignment of Video Embeddings}

\author{%
  Soumyadeep Chandra, and Kaushik Roy \\
  Elmore Family School of Electrical and Computer Engineering\\ Purdue University, West Lafayette, IN 47907, USA \\
  \texttt{\{chand133, kaushik\}@purdue.edu} \\
}

\begin{document}

\maketitle

\begin{abstract}
    Real-world instructional videos are long, noisy, and often contain extended background segments, repeated actions, and execution variability that do not correspond to meaningful procedural steps. We propose \textbf{REMAP}, an unsupervised framework for procedure learning based on \textit{Regularized Fused Partial Gromov-Wasserstein Optimal Transport}. REMAP relaxes balanced transport constraints, allowing non-informative or redundant frames to remain unmatched through partial transport. The formulation jointly models semantic similarity and temporal structure, while incorporating Laplacian-based smoothness and structural regularization to prevent degenerate alignments and reduce background interference.
    We evaluate REMAP on large-scale egocentric and third-person benchmarks. The method consistently outperforms state-of-the-art approaches, achieving up to \textbf{11.6\% (+4.45pp)} F1 and \textbf{19.6\% (+4.73pp)} IoU improvements on EgoProceL, and an average \textbf{41\% (+17.15pp)} F1 gain on ProceL and CrossTask. These results highlight the importance of partial alignment in handling real-world procedural variability and demonstrate that REMAP provides a robust and scalable approach for instructional video understanding.

\end{abstract}

\section{Introduction} \label{intro}
Many emerging AI applications, ranging from assistive robotics to augmented reality systems, depend on the ability to understand and reproduce complex human procedures. Unlike single-step actions, real-world tasks such as cooking, assembling objects, or performing maintenance require learning the underlying procedural structure over an extended temporal structure, where individual steps form a coherent but flexible sequence. Learning such knowledge from visual demonstrations is significantly more challenging than short-term action recognition. Early approaches tried to encode procedures using manually defined rules that specified step boundaries and transitions. While intuitive in constrained environments, these systems struggle to generalize to real-world settings, breaking down under visual variability, background clutter, and execution diversity (e.g., adding dressing before or after chopping vegetables). Real demonstrations are inherently messy: steps may be repeated, reordered, or interleaved with idle segments such as waiting, tool preparation, or camera motion. These challenges have motivated a shift toward learning-based approaches. Unlike action recognition which classifies short clips in isolation (e.g., classifying `cutting' vs. `stirring') ~\cite{carreira2017quo,simonyan2014two,piergiovanni2017learning,kumar2022unsupervised}, \emph{procedure learning (PL)} analyzes collections of demonstrations to automatically discover key procedural steps and their temporal order directly from raw video data, without relying on dense annotations \cite{elhamifar2020self_procel,bansal2022egoprocel_pcass,bansal2024united}. Large collections of uncurated videos, such as online instructional tutorials and egocentric recordings, provide the scale necessary for this task but introduce substantial noise and redundancy \cite{alayrac2016unsupervised,kukleva2019unsupervised}. 

\begin{figure*}[!ht]
  \centering
   \includegraphics[width=\linewidth]{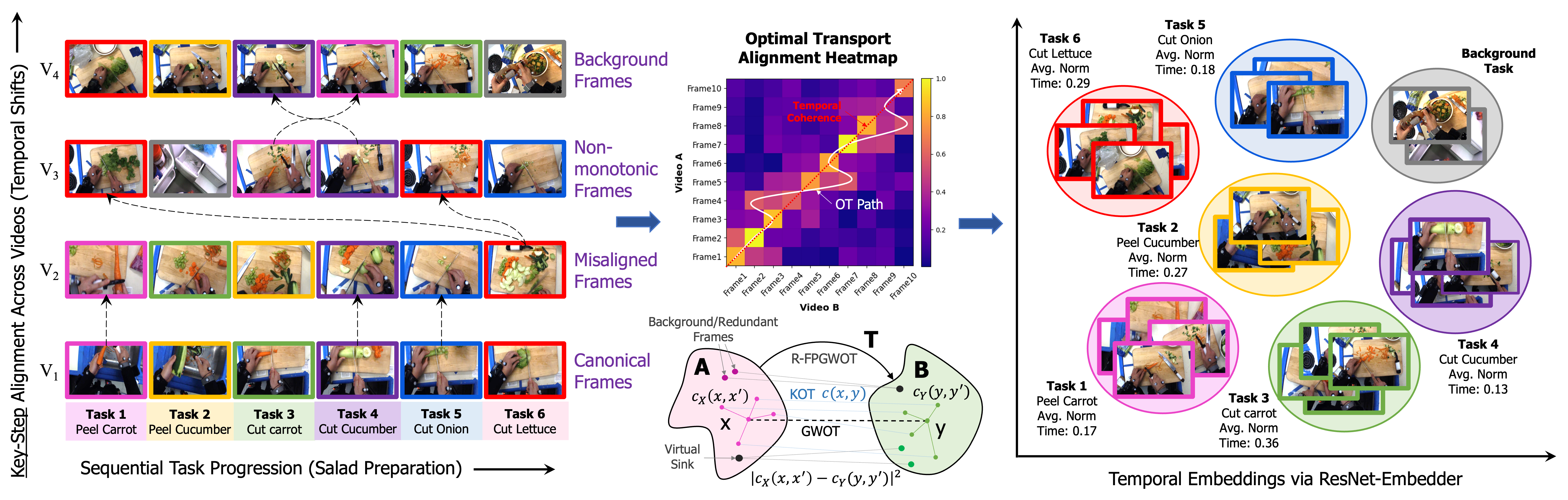}
   \caption{Key-step preparation of a salad bowl \cite{de2009guide_cmummac} illustrating common alignment challenges: (a) \textit{background} frames (gray), (b) \textit{non-monotonic} execution (curved arrows), and (c) \textit{redundant} segments. Two videos are aligned using a transport matrix \textbf{\textit{T}} based on embedding similarity, grouping frames into step-level correspondences (colors). KOT considers only inter-domain costs, whereas GWOT additionally enforces intra-domain structural consistency, yielding smoother temporal mappings. In contrast, the proposed REMAP relaxes balanced-mass constraints via a virtual sink node (black), which absorbs irrelevant or redundant frames (darker shades) and preserves clean, step-wise correspondences across videos.}
  \label{fig:1}
  \vspace{-1em}
\end{figure*}

Prior research has approached PL in supervised and weakly supervised settings. Supervised methods \cite{naing2020procedure,zhou2018towards} require costly frame-level annotations, while weakly supervised approaches \cite{zhukov2019crosstask, li2020set,richard2018neuralnetwork,chang2019d3tw} depend on predefined step lists, limiting scalability. Self-supervised approaches instead exploit temporal consistency (TCC) \cite{dwibedi2019temporal} or cross-video correspondences (CnC) \cite{bansal2022egoprocel_pcass} under largely monotonic alignment assumptions \cite{hadji2021representation}. More recently, optimal transport (OT) has emerged as a powerful framework for aligning instructional videos. Early OT-based methods such as VAVA \cite{shen2021learning} and OPEL \cite{chowdhury2024opel} primarily relied on feature-level matching, while later methods including ASOT \cite{xu2024temporally}, VASOT \cite{ali2025joint}, and RGWOT \cite{mahmood2025procedure} incorporated Gromov-Wasserstein OT \cite{peyre2016gromov} to capture relational temporal structure. 
However, real-world instructional videos, often deviate from these assumptions and exhibit temporal irregularities (Fig.~\ref{fig:1}): (a) \textit{background} frames with irrelevant content (e.g., waiting, idle motion, or showing ingredients), (b) \textit{non-monotonic} sequences where steps occur out of order (e.g., add sauce before chopping all vegetables), and (c) \textit{redundant} segments capturing repeated or unnecessary steps. Existing OT-based PL methods predominantly rely on approximately \emph{balanced transport} objectives, implicitly assuming that most frames admit meaningful cross-video correspondences. As a result, irrelevant or ambiguous segments are frequently forced into incorrect matches, degrading procedural alignment quality.

In this work, we argue that effective procedure learning requires \emph{selective alignment}, where unmatched or semantically irrelevant frames should remain unassigned rather than forced into balanced correspondences. To address this, we introduce \textit{Regularized Fused Partial Gromov-Wasserstein Optimal Transport} (R-FPGWOT), a unified transport formulation that jointly models (i) semantic frame similarity through KOT-style feature matching, (ii) relational temporal structure through GWOT, and (iii) unmatched procedural mass through partial and unbalanced transport with a virtual sink. 
Building on this formulation, we propose \textbf{REMAP} (\textit{Regularized Procedure Alignment with Matching Video Embeddings}), an unsupervised framework that combines fused OT alignment, Laplacian temporal priors, and contrastive regularization in a single objective to produce stable and interpretable transport plans for key-step discovery. Frame embeddings are subsequently segmented using graph-cut clustering~\cite{boykov2002fast} and temporally ordered using Hungarian matching~\cite{bansal2024united}. \textbf{REMAP} achieves an average improvement of \textbf{18.9\% (+7.62pp)} F1 and \textbf{30\% (+7.74pp)} IoU scores across both egocentric (EgoProceL \cite{bansal2022egoprocel_pcass}) and third-person (ProceL \cite{elhamifar2020self_procel}, CrossTask \cite{zhukov2019crosstask}) datasets. In particular, REMAP achieves strong gains on cluttered instructional videos such as CMU-MMAC \cite{de2009guide_cmummac} and EGTEAGAZE+ \cite{li2018eye_egteagaze}, highlighting the importance of partial and selective alignment in realistic procedural settings.

In summary, our main contributions are as follows:
\begin{itemize}[leftmargin=3em]
\item We introduce \textbf{REMAP}, a regularized fused partial Gromov-Wasserstein transport formulation for procedure learning that explicitly models unmatched procedural mass through partial and unbalanced alignment.

\item We show that relaxing strict one-to-one balanced correspondences substantially improves robustness to background clutter, redundant actions, and non-monotonic execution patterns commonly observed in instructional videos.

\item REMAP integrates semantic matching, relational temporal structure, Laplacian priors, and contrastive regularization into a unified OT optimization framework to yield interpretable transport plans for stable key-step discovery.

\item REMAP demonstrates consistent improvements over SOTA baselines, achieving \textbf{11.6\% F1-score (4.45 pp)} and \textbf{19.6\% IoU (4.73 pp)} gains on EgoProceL, with particularly strong performance on cluttered and unconstrained datasets.
\end{itemize}

\vspace{-0.6em}
\section{Related Works}
\vspace{-0.5em}
\paragraph{A. \; Self-Supervised Representation Learning for Videos.}
Self-supervised learning extracts supervisory signals directly from data. Early work in this area focused primarily on images, using tasks such as colorization \cite{larsson2016learning,huang2016connectionist}, jigsaw puzzles \cite{carlucci2019domain,kim2018learning,kim2019self}, rotation prediction \cite{gidaris2018unsupervised,feng2019self}, and clustering \cite{caron2018deep,caron2019unsupervised}. Subsequent video-based approaches exploited temporal structure through frame prediction \cite{ahsan2018discrimnet,diba2019dynamonet,han2019video,srivastava2015unsupervised}, temporal ordering \cite{fernando2017self,lee2017unsupervised,misra2016shuffle,xu2019self}, and motion or speed cues \cite{benaim2020speednet,wang2020self,yao2020video}. While these methods learn general-purpose representations, they do not explicitly address the discovery of procedural structure across demonstrations.

\vspace{-0.5em}
\paragraph{B. \; Representations for Procedure Learning.}
PL focuses on discovering key steps and their temporal organization from multiple demonstrations. Prior work has explored temporal cues \cite{kukleva2019unsupervised, vidalmata2021joint}, attention mechanisms \cite{elhamifar2020self_procel}, and cross-video matching \cite{bansal2022egoprocel_pcass} to derive robust embeddings. Graph-based approaches further cluster semantically similar frames across videos \cite{bansal2024united}, but often rely on preprocessing or heuristic background removal. Multi-modal extensions incorporate narration, text, or gaze information \cite{alayrac2016unsupervised,damen2014you,doughty2020action,fried2020learning,malmaud2015s,yu2014instructional,shah2023steps}, though these signals are frequently noisy, misaligned \cite{elhamifar2020self_procel,elhamifar2019unsupervised}, or computationally expensive. Several methods explicitly model temporal alignment across demonstrations. Classical approaches such as CCA \cite{andrew2013deep} and soft-DTW \cite{haresh2021learning} assume synchronized or monotonic sequences, while TCC \cite{dwibedi2019temporal} and GTCC \cite{donahue2024learning} enforce local cycle-consistency. LAV \cite{haresh2021learning} extends DTW-style global alignment but still relies on monotonic assumptions, limiting robustness.

\vspace{-0.5em}
\paragraph{C. \; Video Alignment with Optimal Transport.}
Optimal transport provides a flexible framework for aligning sequences by matching distributions of frames. Early Kantorovich-based OT methods \cite{thorpe2018introduction} combine feature similarity with temporal priors \cite{shen2021learning,chowdhury2024opel}, but struggle with repeated actions and background clutter. Gromov-Wasserstein OT \cite{peyre2016gromov} addresses reordering by aligning relational structure and has been adopted in ASOT \cite{xu2024temporally}, VASOT \cite{ali2025joint}, and RGWOT \cite{mahmood2025procedure}. However, these methods rely on fully balanced transport, forcing all frames to be matched and resulting in spurious alignments in realistic scenarios. Although ASOT extends this idea with unbalanced transport in AS, it is limited to single-video temporal segmentation and does not address cross-video procedural alignment or shared step discovery required in the PL setting.

\vspace{-0.5em}
\paragraph{D. \; Learning Key-step Ordering.}
Most PL methods assume a fixed or approximately monotonic ordering of the key-steps \cite{elhamifar2019unsupervised,kukleva2019unsupervised,vidalmata2021joint}, or ignore the ordering altogether \cite{elhamifar2020self_procel,shen2021learning}. In contrast, as shown in Figure \ref{fig:1}, real-world procedures often admit multiple valid execution orders. By enabling partial alignment, our approach naturally accommodates such variability while preserving coherent procedural representations. 

\begin{figure*}[!ht]
  \centering
   \includegraphics[width=0.97\linewidth]{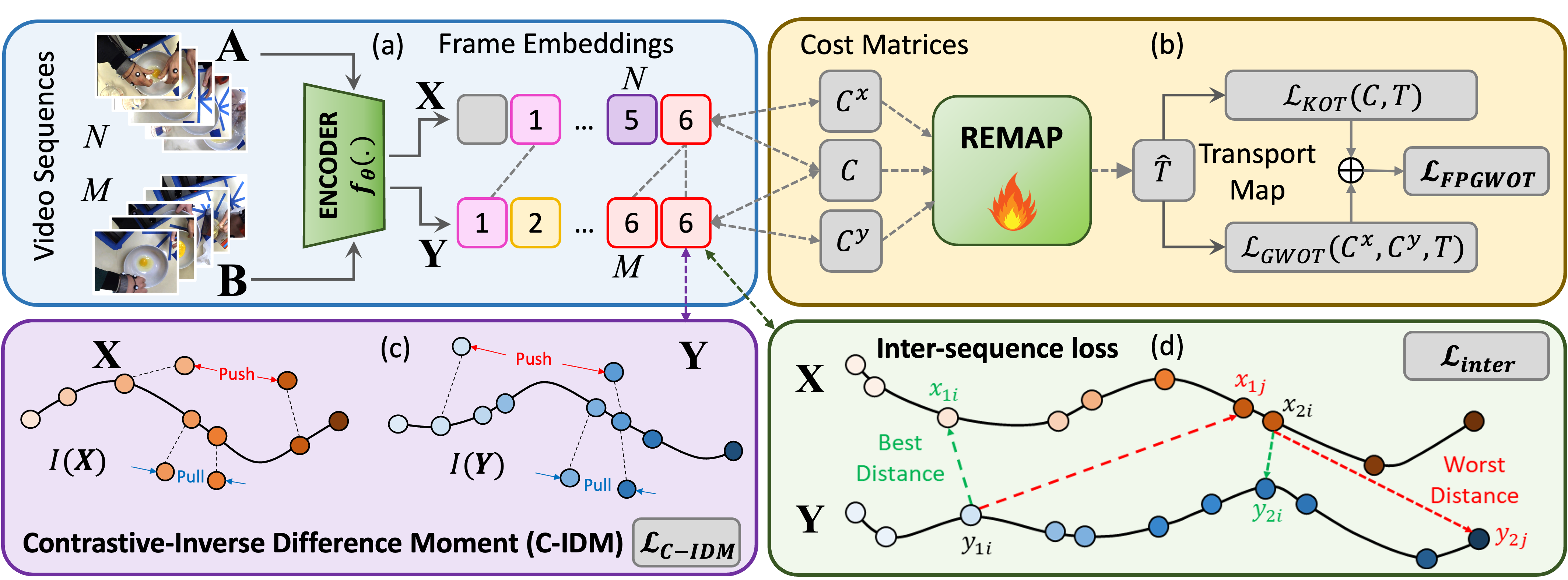}
   \caption{REMAP framework. (a) An encoder extracts frame-level embeddings from two input video sequences for alignment. (b) A fused partial Gromov-Wasserstein optimal transport (FPGWOT) module, guided by structural priors, computes the transport map and establishes frame correspondences. (c) C-IDM pushes apart dissimilar frames while encouraging temporal coherence. (d) An inter-sequence loss stabilizes training by discouraging degenerate alignments and jointly enforcing best and worst match distances. Solid and dashed arrows denote forward and backward computation in temporal alignment, while purple/green highlight regularization components.
   }
  \label{fig:2}
  \vspace{-1.5em}
\end{figure*}

\vspace{-0.6em}
\section{Methodology}
\vspace{-0.6em}
REMAP aligns instructional videos while preserving semantic correspondence and temporal structure, and remains robust to background noise and redundancy. We achieve this by combining partial optimal transport, structural priors, and contrastive regularization into a single training objective. The following subsections describe the OT formulation and the regularizers used to obtain stable transport plans and reliable key-step discovery.

\vspace{-0.4em}
\paragraph{A. \; Regularized Partial Gromov-Wasserstein Optimal Transport (R-FPGWOT) Formulation.}
\vspace{-0.3em}
Optimal Transport (OT) aligns two distributions by moving mass while minimizing a transportation cost \cite{villani2009optimal}. Given two videos $\textbf{A}$ and $\textbf{B}$ with $N$ and $M$ frames, an encoder $f_\theta$ (Fig.~\ref{fig:2}(a)) produces frame embeddings
$\boldsymbol X=[x_1,\ldots,x_N]^\top \in \mathbb R^{N\times D}$ and
$\boldsymbol Y=[y_1,\ldots,y_M]^\top \in \mathbb R^{M\times D}$.
We model each video as an empirical distribution
$\mu=\sum_{i=1}^N \alpha_i \delta_{x_i}$ and
$\nu=\sum_{j=1}^M \beta_j \delta_{y_j}$ with uniform weights
$\alpha_i=\tfrac1N$ and $\beta_j=\tfrac1M$. 

Classical \emph{Kantorovich OT (KOT)} aligns frames using feature similarity, while \emph{Gromov-Wasserstein OT (GWOT)} aligns relational structure within each sequence. Their combination motivates fused GWOT (Fig.~\ref{fig:2}(b)), which improves both semantic fidelity and temporal coherence. However, real instructional videos contain background, idle motion, repeated actions, and skipped steps. Therefore, enforcing a fully balanced frame-to-frame coupling can introduce spurious matches. We address this by extending RGWOT \cite{mahmood2025procedure} with two complementary mechanisms: (i) KL-based marginal relaxation, which permits soft mass imbalance, and (ii) an explicit virtual sink, which provides a finite-cost destination for frames that should remain unmatched.

For the non-augmented coupling $\boldsymbol T\in\mathbb R_+^{N\times M}$, the partial fused objective (FPGWOT) is:
\begin{equation}
\min_{\boldsymbol T\ge 0} 
\underbrace{(1-\rho)\langle \boldsymbol C,\boldsymbol T\rangle}_{\text{KOT}}
+ \underbrace{\rho \sum_{i,k}\sum_{j,l} L(\boldsymbol C^x_{ik},\boldsymbol C^y_{jl})T_{ij}T_{kl}}_{\text{GWOT}}
+ \underbrace{\tau \Big(\mathrm{KL}(\boldsymbol T\mathbf 1\Vert \alpha)
+ \mathrm{KL}(\boldsymbol T^\top \mathbf 1\Vert \beta)\Big)}_{\text{Partial/Unbalanced marginal relaxation}}
-\epsilon h(\boldsymbol T),
\label{eq:FPGWOT_short}
\end{equation}

\vspace{-0.6em}
where $\boldsymbol C_{ij}=\|x_i-y_j\|_2$ is the appearance cost, and
$\boldsymbol C^x\in\mathbb R^{N\times N}$ and
$\boldsymbol C^y\in\mathbb R^{M\times M}$ are positive temporal similarity kernels within videos $\textbf{A}$ and $\textbf{B}$, respectively. Since $\boldsymbol C^x$ and $\boldsymbol C^y$ encode similarity rather than distance, the GW term is written as a reward inside the minimization objective. This encourages matches that preserve temporal-relational similarity instead of penalizing them. $T_{ij}$ reflects how much mass of frame $x_i$ is transported to frame $y_j$. The coefficient $\tau>0$ controls how strongly the relaxed marginals are enforced. Entropic regularization $-\epsilon h(\boldsymbol T)$ \cite{peyre2016gromov,cuturi2013sinkhorn}, where
$h(\boldsymbol T)=-\sum_{i,j}t_{ij}\log t_{ij}$, improves numerical stability and enables efficient Sinkhorn-style updates.

\begin{figure}[!ht]
   \vspace{-0.4em}
   \centering
   \includegraphics[width=0.98\linewidth]{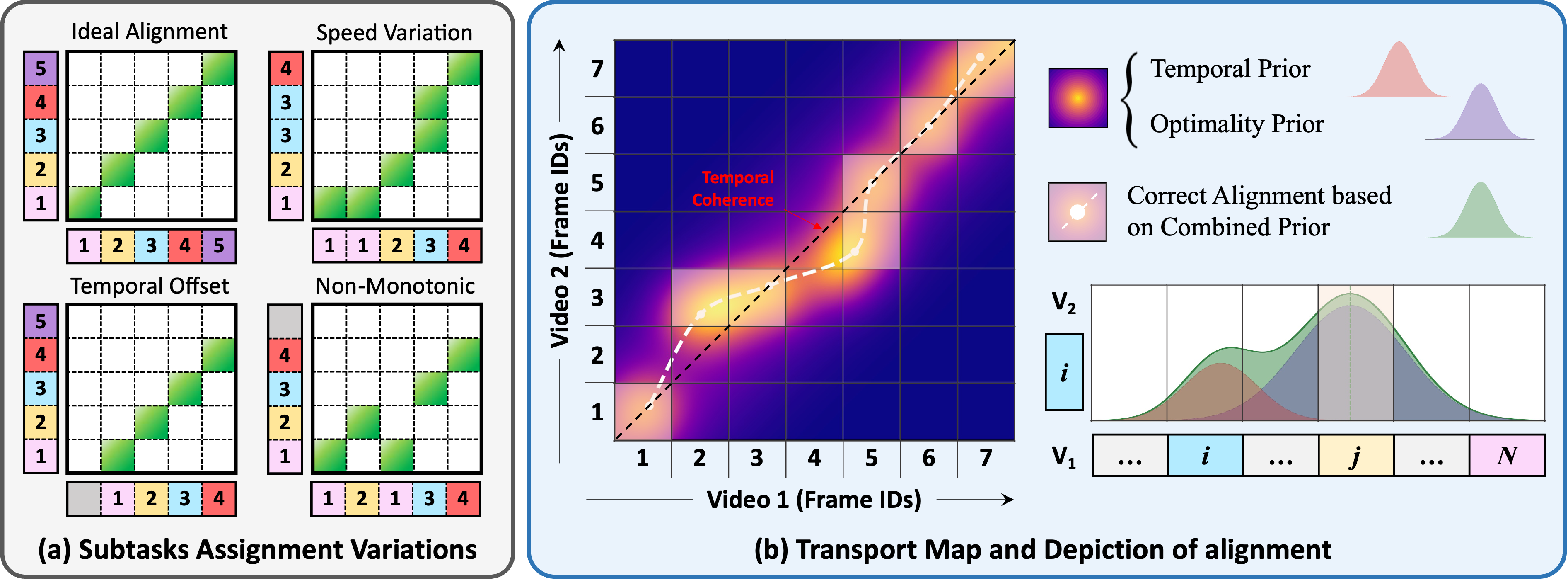}
   \vspace{0.1em}
   \caption{(a) Examples of pairwise alignment scenarios captured by the assignment matrix. (b) Visualization of the OT map in 2D, along with a 1D illustration showing how $i$-th frame from Video 2 aligns with its best-matched $j$-th frame from Video 1.
   }
  \label{fig:3}
  \vspace{-1.5em}
\end{figure}

\vspace{-0.2em}
\paragraph{B. \; Regularization using Priors.}
GWOT aligns sequences by matching pairwise relational structure. Under repeated actions, camera motion, or clutter, the learned transport plan may still assign semantically similar but temporally distant frames to each other. We therefore introduce Laplace-shaped \emph{Temporal} and \emph{Optimality} priors \cite{liu2022learning,chowdhury2024opel} that softly bias transport mass toward plausible near-diagonal alignments while allowing early starts, speed variations, and moderate non-monotonic deviations (Fig.~\ref{fig:3}). For real frame indices $i\le N$ and $j\le M$, the prior is:

\vspace{-0.5em}
\begin{equation}
\begin{alignedat}{2}
\boldsymbol Q(i,j)
& = \phi\exp\Big(\frac{-|d_t(i,j)|}{b}\Big)
 + (1-\phi)\exp\Big(\frac{-|d_o(i,j)|}{b}\Big) \\
\text{where,} \quad d_t(i,j)
& = \frac{\big|\,i/N - j/M\,\big|}{\sqrt{1/N^2 + 1/M^2}}, 
\quad d_o(i,j)
= \frac{\big|\,i/N - i_o/N\,\big| + \big|\,j/M - j_o/M\,\big|}
{2\sqrt{1/N^2 + 1/M^2}}
\end{alignedat}
\label{eq:mixprior_short}
\end{equation}

\vspace{-0.5em}
Here, $d_t(i,j)$ preserves global temporal order, while $d_o(i,j)$ measures proximity to an estimated alignment center $(i_o,j_o)$. In practice, $(i_o,j_o)$ is computed from the current best appearance-based correspondence, and $\phi$ is annealed from $1$ to $0.5$ during training as seen in VAVA \cite{liu2022learning}. This begins with a strong temporal prior and gradually permits a more flexible non-monotonic alignment.

\vspace{-0.5em}
\paragraph{Virtual frame for partial alignment.}
To explicitly absorb background or redundant frames, we augment the coupling with a virtual frame on both axes,
$\hat{\boldsymbol T}\in\mathbb R_+^{(N+1)\times(M+1)}$. The last row and last column correspond to the sink. The augmented appearance cost and temporal structure matrices are:
\vspace{-0.4em}
\begin{equation}
\hat{\boldsymbol C}=
\begin{bmatrix}
\boldsymbol C & \zeta\mathbf 1_N\\
\zeta\mathbf 1_M^\top & 0
\end{bmatrix},
\qquad
\hat{\boldsymbol C}^x=
\begin{bmatrix}
\boldsymbol C^x & \boldsymbol 0\\
\boldsymbol 0^\top & 0
\end{bmatrix},
\qquad
\hat{\boldsymbol C}^y=
\begin{bmatrix}
\boldsymbol C^y & \boldsymbol 0\\
\boldsymbol 0^\top & 0
\end{bmatrix}.
\label{eq:sink_cost}
\end{equation}

\vspace{-1em}
where $\zeta$ is the finite virtual-frame threshold. A smaller $\zeta$ makes sink assignment easier, while a larger $\zeta$ encourages real-frame matching unless the match is unreliable. Thus, the virtual frame is masked from temporal-distance computations and remains position-invariant. The augmented prior $\hat{\boldsymbol Q}$ copies $\boldsymbol Q$ on real-frame entries and assigns constant positive values to sink-related entries:
\vspace{-0.4em}
\begin{equation}
\hat Q_{ij}=Q_{ij}\;\;(i\le N,j\le M),\quad
\hat Q_{i,M+1}=q_{\mathrm{sink}},\quad
\hat Q_{N+1,j}=q_{\mathrm{sink}},\quad
\hat Q_{N+1,M+1}=q_{\mathrm{ss}}.
\end{equation}

\vspace{-1em}
The augmented marginals $\hat{\boldsymbol\alpha}$ and $\hat{\boldsymbol\beta}$ include a small positive sink mass, ensuring that all entries remain numerically well-defined.

\vspace{-0.4em}
\paragraph{C. \; IDM-style Structural Regularization with FPGWOT.}
We regularize $\hat{\boldsymbol T}$ using inverse-distance moments (IDM) \cite{albregtsen2008statistical,liu2022learning}. The first term promotes diagonal concentration, while the second encourages sharp alignment ridges around the estimated correspondence center:
\vspace{-0.2em}
\begin{equation}
\begin{aligned}
M(\hat{\boldsymbol T})=\phi \sum_{ij}\frac{t_{ij}}{(\tfrac{i}{N+1}-\tfrac{j}{M+1})^2+1} 
+ (1-\phi)\sum_{ij}\frac{t_{ij}}{\tfrac{1}{2}d_m+1},
\; \quad d_m=\Big(\frac{i-i_o}{N+1}\Big)^2+\Big(\frac{j-j_o}{M+1}\Big)^2
\label{eq:Mcomb}
\end{aligned}
\end{equation}

\vspace{-1em}
The sums are taken only over real frame indices. Sink entries are excluded from $M(\hat{\boldsymbol T})$, preventing the virtual frame from receiving any artificial temporal position bias.

\vspace{-0.5em}
\paragraph{Constrained formulation.}
We embed these priors into a constrained feasible set of the FPGWOT formulation. Unlike balanced OT, our formulation permits soft mass imbalance while constraining the structure of $\hat{\boldsymbol T}$ \cite{xu2024temporally,bai2025fused}. Specifically, we require (i) a sufficiently high structural score $M(\hat{\boldsymbol T})\ge\xi_1$, and (ii) proximity to the augmented prior $\hat{\boldsymbol Q}$ through $\mathrm{KL}(\hat{\boldsymbol T}\Vert\hat{\boldsymbol Q})\le\xi_2$:
\vspace{-0.4em}
\begin{equation}
U_{\xi_1,\xi_2}(\hat{\boldsymbol\alpha},\hat{\boldsymbol\beta})
=
\{ \hat{\boldsymbol T} \ge 0 :\,
\hat{\boldsymbol T}\boldsymbol 1_{M+1} \approx \hat{\boldsymbol\alpha},\quad
\hat{\boldsymbol T}^\top \boldsymbol 1_{N+1} \approx \hat{\boldsymbol\beta},M(\hat{\boldsymbol T}) \ge \xi_1,\quad
\mathrm{KL}(\hat{\boldsymbol T}\Vert\hat{\boldsymbol Q}) \le \xi_2
\}.
\label{eq:feasset}
\end{equation}

\vspace{-1em}
Introducing Lagrange multipliers $\lambda_1,\lambda_2>0$ gives the following regularized inner objective after linearizing the structural GW reward at iteration $s$:
\vspace{-0.6em}
\begin{equation}
\hat{\boldsymbol T}^{\text{\scriptsize (\textit{s+1})}}
=
\arg\min_{\hat{\boldsymbol T}\ge 0}
\Big\langle \hat{\boldsymbol T},\,\widetilde{\boldsymbol D}^{\text{\scriptsize (\textit{s})}}\Big\rangle
-\lambda_1 M(\hat{\boldsymbol T})
+\lambda_2\,\mathrm{KL}\Big(\hat{\boldsymbol T}\Vert\hat{\boldsymbol Q}^{\text{\scriptsize (\textit{s})}}\Big)
+\tau\Big(
\mathrm{KL}(\hat{\boldsymbol T}\mathbf{1}_{M+1}\Vert \hat{\boldsymbol\alpha})
+\mathrm{KL}(\hat{\boldsymbol T}^\top\mathbf{1}_{N+1}\Vert \hat{\boldsymbol\beta})
\Big),
\label{eq:dual}
\end{equation}

\vspace{-1em}
where, $ \widetilde{\boldsymbol D}^{\text{\scriptsize (\textit{s})}} = (1-\rho)\hat{\boldsymbol C} -\rho\,\boldsymbol (2 \hat{\boldsymbol C}^x\hat{\boldsymbol T}^{(s)}\hat{\boldsymbol C}^y)$.
The negative sign before $\rho$ reflects that $\hat{\boldsymbol C}^x$ and $\hat{\boldsymbol C}^y$ are similarity kernels; therefore, structurally consistent matches reduce the effective transport cost. 

The solution of Eq.~\ref{eq:dual} has the Sinkhorn form:
\vspace{-0.4em}
\begin{equation}
\hat{\boldsymbol T}^{\text{\scriptsize (\textit{s+1})}}
=
\mathrm{Diag}(u^{\text{\scriptsize (\textit{s})}})\boldsymbol K^{\text{\scriptsize (\textit{s})}}\mathrm{Diag}(v^{\text{\scriptsize (\textit{s})}}),
\quad
\text{with Gibbs kernel:} \;
K_{ij}^{\text{\scriptsize (\textit{s})}}
=
\hat Q_{ij}^{\text{\scriptsize (\textit{s})}}
\exp\left(
\frac{
\lambda_1 s_{ij}^{\text{\scriptsize (\textit{s})}}
-\widetilde D_{ij}^{\text{\scriptsize (\textit{s})}}
}{\lambda_2}
\right),
\label{eq:gibbs_kernel_main}
\end{equation}

\vspace{-0.6em}
where the IDM score is:
\vspace{-0.4em}
\begin{equation}
s_{ij}^{(s)}
=
\left[
\phi^{(s)}
\frac{1}{(\tfrac{i}{N+1}-\tfrac{j}{M+1})^2+1}
+
(1-\phi^{(s)})
\frac{1}{\tfrac12 d_m(i,j)+1}
\right].
\label{eq:idm_score_fixed}
\end{equation}

\vspace{-0.6em}
At each outer iteration, $\phi^{(s)}$ is annealed and both $\hat{\boldsymbol Q}^{(s)}$ and $s_{ij}^{\lambda_1,(s)}$ are recomputed. Thus, the annealing schedule directly affects the transport update. The resulting KL-regularized subproblem is solved with unbalanced Sinkhorn updates, and the overall procedure converges to a stationary point under the standard assumptions used in entropic GW optimization. Details are provided in Appendix~\ref{app:unb}.

\vspace{-0.6em}
\paragraph{D. \; Contrastive stabilization.}
To avoid degenerate or collapsed alignments, we incorporate two contrastive regularizers. The \emph{intra-sequence} C-IDM loss \cite{haresh2021learning,liu2022learning} (Eq.~\ref{eq:cidm}) enforces temporal smoothness by pulling nearby frames together while pushing distant frames apart
(Fig. \ref{fig:2}(c)).
\begin{equation}
\begin{aligned}
& I(\boldsymbol X) =\sum_{i,j}\mathcal N(i,j)\frac{d(i,j)}{\gamma(i,j)} +\big(1-\mathcal N(i,j)\big)\gamma(i,j)\max\{0,\lambda_3-d(i,j)\}
\\ \text{where, } &\quad \mathcal N(i,j)=\mathbf{1}\{|i-j|\le\delta\}, \gamma(i,j)=(i-j)^2+1, \quad d(i,j)=\|\boldsymbol x_i-\boldsymbol x_j\|_2.
\label{eq:cidm}
\end{aligned}
\end{equation}

\vspace{-0.2em}
The \emph{inter-sequence} contrastive loss \cite{chowdhury2024opel} (Eq.~\ref{eq:inter_fpgwot}) further stabilizes training by separating best and worst cross-video matches according to the learned transport plan(Fig \ref{fig:2}(d)). 
\vspace{-0.2em}
\begin{equation}
\mathcal L_{\text{inter}} =
\mathrm{CE}\!\left(
\begin{bmatrix}
\text{best\_dist}\\[2pt]
\text{worst\_dist}
\end{bmatrix},
\begin{bmatrix}
0\\[2pt]
1
\end{bmatrix}
\right),
\text{(best/worst) from arg max/min of }\hat{\boldsymbol T}^{\text{R-FPGW}}_{\lambda_1,\lambda_2}\text{ along rows/cols}.
\label{eq:inter_fpgwot} 
\end{equation}

Together, these losses encourage discriminative yet temporally coherent embeddings. The final REMAP objective combines the R-FPGWOT alignment loss (Eq.~\ref{eq:dual}) with intra- and inter-sequence contrastive regularization, preserving both embedding diversity and cross-video separability.
\vspace{-0.2em}
\begin{equation}
\mathcal L_{\text{REMAP}}
=
c_1\,\mathcal L_{\text{R-FPGWOT}}
+c_2\mathcal L_{\text{C-IDM}}
+c_3\,\mathcal L_{\text{inter}}=
c_1\,\ell^{\text{R-FPGW}}_{\lambda_1,\lambda_2,\tau}(\boldsymbol X,\boldsymbol Y)
+c_2\big(I(\boldsymbol X)+I(\boldsymbol Y)\big)
+c_3\,\mathcal L_{\text{inter}}
\label{eq:final_fpgwot}
\end{equation}


\vspace{-0.4em}
\paragraph{E. \; Clustering and Key-Step Ordering.}
After OT training, we localize key steps using post-hoc multi-label graph-cut segmentation \cite{greig1989exact} with temporal smoothness. We assign each frame into one of the $K$ clusters as defined in Eq.~\ref{eqn:graphcut}. We further order based on average normalized timestamps to capture the procedural structure \cite{bansal2022egoprocel_pcass,chowdhury2024opel}. The alignment is entirely determined by the differentiable OT-based objective. Graph-cut stage is applied \emph{after} the transport matrix $\boldsymbol T$ is fully learned and does \emph{not} influence the OT optimization. It converts the learned embeddings into contiguous temporal segments. Details are provided in Appendix~\ref{app8.1} and supported by ablations in Table~\ref{tab:table_cluster}, indicating that graph-cut yields cleaner segment boundaries without altering the fundamental behavior of the learned alignment model. 

\vspace{-0.4em}
\section{Experiments and Results}
\label{results}
\vspace{-0.5em}
\paragraph{Datasets.}
We evaluate REMAP across both egocentric and third-person perspectives. For third-person analysis, we use \textbf{ProceL} \cite{elhamifar2020self_procel}, with 720 videos spanning 12 tasks over 47.3 hours and \textbf{CrossTask} \cite{zhukov2019crosstask}, which contains 2763 videos (213 hours) covering 18 primary tasks. For egocentric evaluation, we adopt the large-scale \textbf{EgoProceL} benchmark \cite{bansal2022egoprocel_pcass}, featuring 62 hours of head-mounted recordings from 130 users performing 16 subtasks. Dataset statistics are summarized in Appendix Table~\ref{tab:abs2}.

\vspace{-0.8em}
\paragraph{Evaluation.}
We follow the evaluation practices of recent OT-based procedure learning methods \cite{chowdhury2024opel,mahmood2025procedure}, reporting both F1-score and temporal Intersection-over-Union (IoU). Framewise scores are computed per key-step and averaged across steps. Precision measures the proportion of correctly predicted key-step frames among all predicted frames, while recall measures the proportion of ground-truth key-step frames correctly retrieved. Following prior works \cite{bansal2022egoprocel_pcass,elhamifar2020self_procel,shen2021learning}, the Hungarian algorithm \cite{kuhn1955hungarian} is used to establish a one-to-one mapping between predicted and ground-truth steps.

\vspace{-0.8em}
\paragraph{Experimental Setup.}
We use a ResNet-50 backbone pretrained on ImageNet for frame-level feature extraction, following \cite{bansal2022egoprocel_pcass,chowdhury2024opel}. Features are extracted from the Conv4c layer and concatenated using a two-frame temporal window. These are processed by two 3D convolution layers, followed by global max pooling and two fully connected layers, producing 128-dimensional embeddings. The encoder is trained on video pairs with random temporal sampling using the proposed $\mathcal L_{\text{REMAP}}$ objective. 

\begin{table}[!ht]
\centering
\vspace{-1em}
\caption{Results on EgoProceL comparing REMAP with OT-based and prior baselines. Best and second-best scores are in bold and underlined. STEPS \cite{shah2023steps} (\textcolor{Purple}{purple}) uses extra modalities (flow, gaze, depth), while our method relies only on visuals. OT-based SOTA methods are shown in \textcolor{gray}{gray}, and our work REMAP is highlighted in \textcolor{cyan}{blue}.}
\label{tab:table1}
\vspace{0.1em}
\resizebox{\linewidth}{!}{%
\begin{tabular}{>{\rule[-0.1cm]{0pt}{0.41cm}}lccccccccccccccccc}
\toprule
 & \multicolumn{17}{c}{EgoProceL} \\ \cline{2-18} 
 & \multicolumn{2}{c}{CMU-MMAC} &  & \multicolumn{2}{c}{EGTEA-GAZE+} &  & \multicolumn{2}{c}{MECCANO} &  & \multicolumn{2}{c}{EPIC-Tents} &  & \multicolumn{2}{c}{PC Assembly} &  & \multicolumn{2}{c}{PC Disassembly} \\ \cline{2-3} \cline{5-6} \cline{8-9} \cline{11-12} \cline{14-15} \cline{17-18} 
 & F1 & IoU &  & F1 & IoU &  & F1 & IoU &  & F1 & IoU &  & F1 & IoU &  & F1 & IoU \\ \hline
Random & 15.7 & 5.9 &  & 15.3 & 4.6 &  & 13.4 & 5.3 &  & 14.1 & 6.5 &  & 15.1 & 7.2 &  & 15.3 & 7.1 \\
Uniform & 18.4 & 6.1 &  & 20.1 & 6.6 &  & 16.2 & 6.7 &  & 16.2 & 7.9 &  & 17.4 & 8.9 &  & 18.1 & 9.1 \\
CnC \cite{bansal2022egoprocel_pcass} & 22.7 & 11.1 &  & 21.7 & 9.5 &  & 18.1 & 7.8 &  & 17.2 & 8.3 &  & 25.1 & 12.8 &  & 27.0 & 14.8 \\
UG-I3D \cite{bansal2024united} & 28.4 & 15.6 &  & 25.3 & 14.7 &  & 18.3 & 8.0 &  & 16.8 & 8.2 &  & 22.0 & 11.7 &  & 24.2 & 13.8 \\
GPL-w BG \cite{bansal2024united} & 30.2 & 16.7 &  & 23.6 & 14.9 &  & 20.6 & 9.8 &  & 18.3 & 8.5 &  & 27.6 & 14.4 &  & 26.9 & 15.0 \\
GPL-w/o BG \cite{bansal2024united} & 31.7 & 17.9 &  & 27.1 & 16.0 &  & 20.7 & 10.0 &  & 19.8 & 9.1 &  & 27.5 & 15.2 &  & 26.7 & 15.2 \\
\rowcolor{Purple!20}STEPS \cite{shah2023steps} & 28.3 & 11.4 &  & 30.8 & 12.4 &  & 36.4 & 18.0 &  & \textbf{42.2} & 21.4 &  & 24.9 & 15.4 &  & 25.9 & 14.6 \\
\rowcolor{lightgray!35} OPEL \cite{chowdhury2024opel} & 36.5 & 18.8 &  & 29.5 & 13.2 &  & 39.2 & 20.2 &  & 20.7 & 10.6 &  & 33.7 & 17.9 &  & 32.2 & 16.9 \\ 
\rowcolor{lightgray!35} RGWOT \cite{mahmood2025procedure} & 54.4 & 38.6 &  & 37.4 & 22.9 &  & \underline{59.5} & \underline{42.7} &  & 39.7 & 24.9 &  & \textbf{43.6} & \textbf{28.0} &  & \textbf{45.9} & \textbf{30.1} \\
\rowcolor{cyan!15} \textit{\textbf{REMAP}} (R-FGWOT) (Ours) & \underline{58.3} & \underline{42.5} &  & \underline{62.4} & \underline{47.2} &  & 59.1 & 42.3 &  & 39.1 & \underline{24.4} &  & 40.9 & 25.4 &  & 41.9 & 28.1 \\
\rowcolor{cyan!15} \textit{\textbf{REMAP}} (R-FPGWOT) (Ours) & \textbf{59.7} & \textbf{43.7} &  & \textbf{64.2} & \textbf{49.3} &  & \textbf{59.6} &  \textbf{42.7} &  & \underline{39.8} & \textbf{25.0} &  & \underline{41.4} & \underline{26.3} &  & \underline{42.5} & \underline{28.6} \\
\hline
\bottomrule
\end{tabular}%
}
\vspace{-1.6em}
\end{table} 

REMAP is controlled by three interpretable parameters: the partial marginal relaxation weight $\tau$, the KOT--GWOT trade-off $\rho$, and the Laplace scale $b$. Unless otherwise stated, we use a fixed global configuration across datasets and report sensitivity analyses in Appendix~\ref{app8.2}. Remaining weights $(\lambda_1,\lambda_2,c_1,c_2,c_3)$ are fixed across datasets as listed in Table~\ref{tab:hyper-param}. To reduce randomness, all results are averaged over three independent runs. We additionally report variance statistics in the App.~\ref{app8.5}.

\vspace{-0.8em}
\paragraph{Results on Egocentric View.}
Table~\ref{tab:table1} compares REMAP with prior PL and OT-based baselines on EgoProceL~\cite{bansal2022egoprocel_pcass}. REMAP achieves the best performance on CMU-MMAC, EGTEA-GAZE+, MECCANO, and EPIC-Tents, while remaining competitive on PC Assembly and PC Disassembly. The largest gains occur on datasets containing substantial background or redundant content. For example, REMAP improves over RGWOT\cite{mahmood2025procedure} by $\textbf{+26.8}$ F1 points on EGTEA-GAZE+ and $\textbf{+5.3}$ F1 points on CMU-MMAC. Partial alignment proves beneficial when high levels of clutter and temporal variability prevent consistent cross-video mapping across all frames. Overall, REMAP achieves average relative improvements of \textbf{11.6\% F1 (+4.45pp)} and \textbf{19.6\% IoU (+4.73pp)} across EgoProceL.

\begin{wraptable}{r}{0.55\columnwidth}
\centering
\vspace{-1.2em}
\caption{PL results on third-person datasets. P (Precision), R (Recall), and F1-score. The \textbf{best} and \underline{second-best} results are highlighted.}
\label{tab:table2}
\vspace{-0.1em}
\resizebox{0.54\columnwidth}{!}{%
\begin{tabular}{>{\rule[-0.1cm]{0pt}{0.4cm}}lccccccc}
\toprule
 & \multicolumn{3}{c}{ProceL} &  & \multicolumn{3}{c}{CrossTask} \\ \cline{2-4} \cline{6-8} 
 & P & R & F1 &  & P & R & F1 \\ 
\hline 
Uniform & 12.4 & 9.4 & 10.3 &  & 8.7 & 9.8 & 9.0 \\
\rowcolor{Goldenrod!25}Alayrac et al. \cite{alayrac2016unsupervised} & 12.3 & 3.7 & 5.5 &  & 6.8 & 3.4 & 4.5 \\
\rowcolor{Goldenrod!25}Kukleva et al. \cite{kukleva2019unsupervised} & 11.7 & 30.2 & 16.4 &  & 9.8 & 35.9 & 15.3 \\
Elhamifar \& Huynh \cite{elhamifar2020self_procel} & 9.5 & 26.7 & 14.0 &  & 10.1 & 41.6 & 16.3 \\
Shen et al. \cite{shen2021learning} & 16.5 & 31.8 & 21.1 &  & 15.2 &  35.5 & 21.0 \\
CnC \cite{bansal2022egoprocel_pcass} & 20.7 & 22.6 & 21.6 &  & 22.8 & 22.5 & 22.6 \\
UG-I3D \cite{bansal2024united} & 21.3 & 23.0 & 22.1 &  & 23.4 & 23.0 & 23.2 \\
GPL \cite{bansal2024united} & 22.4 & 24.5 & 23.4 &  & 24.9 & 24.1 & 24.5 \\
\rowcolor{Purple!15}STEPS \cite{shah2023steps} & 23.5 & 26.7 & 24.9 &  & 26.2 & 25.8 & 25.9 \\
\rowcolor{lightgray!30} OPEL \cite{chowdhury2024opel} & 33.6 & 36.3 & 34.9 &  & 35.6 & 34.8 & 35.1 \\ 
\rowcolor{lightgray!30} RGWOT \cite{mahmood2025procedure} & 42.2 & 46.7 & 44.3 &  & 40.4 & 40.7 & 40.4 \\
\rowcolor{cyan!10} \textit{\textbf{REMAP}} (R-FGWOT) & \underline{53.5} & \underline{60.4} & \underline{56.7} &  & \underline{60.2} & \underline{61.2} & \underline{60.6} \\
\rowcolor{cyan!10} \textit{\textbf{REMAP}} (R-FPGWOT) & \textbf{54.4} & \textbf{61.5} & \textbf{57.6} &  & \textbf{60.9} & \textbf{61.9} & \textbf{61.4} \\
\hline
\bottomrule
\end{tabular}%
}
\vspace{-1em}
\end{wraptable}

\vspace{-1em}
\paragraph{Results on Third-person View.}
We further evaluate REMAP on ProceL~\cite{elhamifar2020self_procel} and CrossTask~\cite{zhukov2019crosstask} in Table~\ref{tab:table2}. REMAP substantially improves over RGWOT, achieving \textbf{57.6} F1 on ProceL and \textbf{61.4} F1 on CrossTask. This corresponds to absolute gains of \textbf{+13.3} and \textbf{+21.0} F1 points, respectively. The improvement is consistent across precision and recall, indicating that REMAP does not simply collapse into a single degenerate solution~\cite{kukleva2019unsupervised,elhamifar2020self_procel}, but rather better separates meaningful procedural frames from background or ambiguous segments. Detailed subtask-level results are reported in App. Tables~\ref{tab:CMU-3rdperson} and~\ref{tab:abs5}.

\vspace{-1em}
\paragraph{Comparison with Multimodal Methods.}
We additionally compare REMAP with multimodal PL methods that use richer modalities such as gaze, depth, optical flow, or narration in Table~\ref{tab:table1},\ref{tab:table2}. \textcolor{Purple}{STEPS}~\cite{shah2023steps} combines RGB with depth and gaze information, while REMAP relies only on RGB inputs. Despite this limitation, REMAP outperforms STEPS on most EgoProceL datasets and achieves stronger IoU on EPIC-Tents~\cite{jang2019epictents}, where STEPS attains slightly higher F1. On third-person datasets, REMAP also surpasses \textcolor{Dandelion}{narration-based} methods~\cite{alayrac2016unsupervised,shen2021learning}. These results suggest that structured partial transport provides strong procedural supervision even without auxiliary modalities.

\begin{figure}[!ht]
  \centering
   \vspace{-0.6em}
   \includegraphics[width=0.98\linewidth]{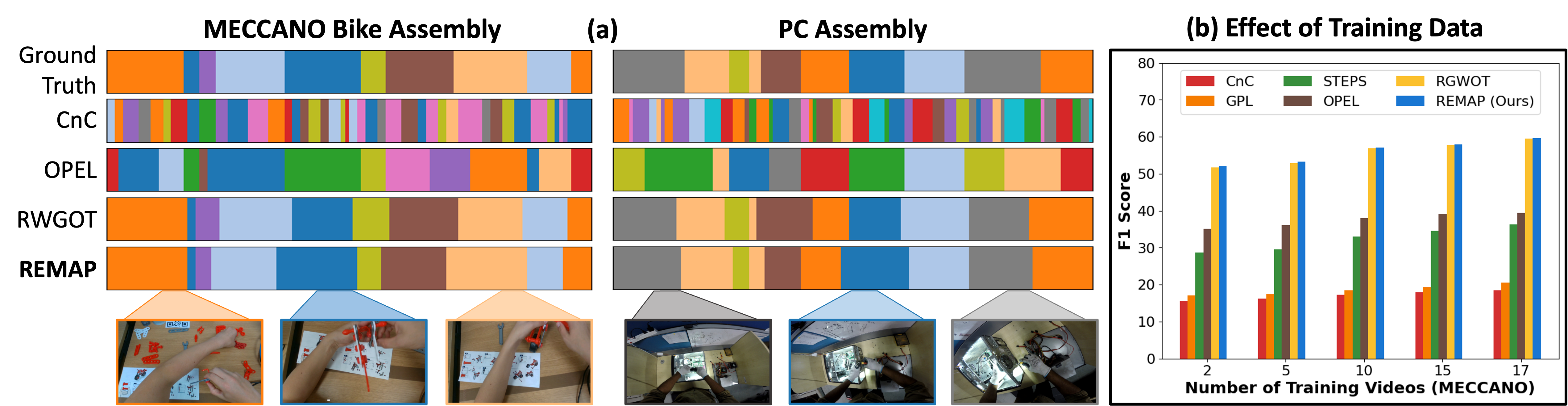}
   \vspace{-0.4em}
   \caption{(a) Qualitative outcomes on MECCANO and PC Assembly, where color highlights distinguish sub-tasks across key-steps. REMAP achieves superior alignment compared to existing SOTA methods by introducing a virtual frame to effectively manage unmatched frames. (b) Influence of training data volume on encoder performance, highlighting REMAP's data efficiency.}
  \label{fig:4}
  \vspace{-1.4em}
\end{figure}

\vspace{-1em}
\paragraph{Qualitative Results.}
Fig.~\ref{fig:4}(a) compares REMAP with representative baselines. CnC~\cite{bansal2022egoprocel_pcass} often fragments sequences into noisy segments, while OPEL~\cite{chowdhury2024opel} and RGWOT~\cite{mahmood2025procedure} continue to align redundant or background frames to procedural steps. In contrast, REMAP assigns low-confidence regions to the virtual sink and preserves cleaner correspondences among salient procedural frames. This results in more contiguous key-step localization and more interpretable transport maps.

\section{Ablation Study}
\label{ablation}
\vspace{-0.8em}
\paragraph{Component-wise Analysis of the REMAP Loss.} 
Table~\ref{tab:fpgwot_ablation} analyzes the contribution of each component within REMAP, revealing that while structural and temporal priors establish the foundation, the \textbf{partial penalty} ($\tau$) provides critical adaptability.
Adding the GWOT structural term produces the largest improvement over the OPEL-style baseline~\cite{chowdhury2024opel}, increasing F1 by $+12.6$ on MECCANO and $+14.0$ on CMU-MMAC. Temporal Laplace priors~\cite{mahmood2025procedure} further refine alignment by reducing off-diagonal drift and improving temporal coherence. The effect of the partial penalty $\tau$ depends strongly on dataset characteristics: it provides a substantial gain in F1 on CMU-MMAC ($\textbf{+5.3pp}$) but a smaller isolated improvement on MECCANO ($\textbf{+0.5pp}$). This suggests that partial marginal relaxation is most beneficial in cluttered settings, where many frames do not admit reliable correspondences. In cleaner datasets such as MECCANO, structural priors already capture much of the alignment structure, making the additional benefit of partial transport more modest. The contrastive regularizers and KL-based priors improve embedding stability and help prevent degenerate alignments, while contributing consistent gains when combined with the core transport objective.

\begin{table}[!ht]
\centering
\vspace{-1.4em}
\caption{Ablation study of \textbf{REMAP} loss components. We highlight the effects of the GWOT term ($\rho$), Laplace prior ($b$), and partial penalty ($\tau$). \textcolor{ForestGreen}{Green values} denote gains over the closest setting.}
\vspace{0.2em}
\label{tab:fpgwot_ablation}
\resizebox{\linewidth}{!}{%
\begin{tabular}{>{\rule[-0.1cm]{0pt}{0.44cm}}lccccc|cc|cc}
\toprule
\textbf{Configuration} & \textbf{Intra C-IDM} & \textbf{Laplace} & \textbf{Virtual} & \textbf{GWOT} & \textbf{Partial} & \multicolumn{2}{c|}{\textbf{MECCANO}} & \multicolumn{2}{c}{\textbf{CMU-MMAC}} \\
\cline{7-10}
& \textbf{+ Inter-CL} & \textbf{Prior ($b$)} & \textbf{Sink} & \textbf{($\rho$)} & \textbf{Penalty ($\tau$)} & \textbf{F1} & \textbf{IoU} & \textbf{F1} & \textbf{IoU} \\
\midrule

\multicolumn{10}{l}{\textit{Ablation of non-structural components}} \\
\hline

w/o Laplace Prior 
& \checkmark &  & \checkmark &  & 
& 35.8 & 16.1 & 32.6 & 14.4 \\

w/o Contrastive Regularizers 
&  & T+O & \checkmark &  & 
& 36.8 & 17.1 & 36.1 & 16.5 \\

w/o Virtual Frame 
& \checkmark & T+O &  &  & 
& 38.6 & 19.6 & 33.8 & 16.4 \\

\rowcolor{lightgray!35}
\textsuperscript{$\dagger$}\textit{\textbf{OPEL}} baseline
& \checkmark & T+O & \checkmark &  & 
& 39.2 & 20.2 
& 36.5 & 18.8 \\

\hline
\multicolumn{10}{l}{\textit{Effect of main REMAP components}} \\
\hline

+ GWOT structural term ($\rho$)
& \checkmark &  & \checkmark & \checkmark & 
& 51.8 \gain{+12.6} & 35.5 \gain{+15.3}
& 50.5 \gain{+14.0} & 33.7 \gain{+14.9} \\

\rowcolor{lightgray!35}
\textsuperscript{$*$}\textit{\textbf{RGWOT} baseline}: + Temporal Laplace prior ($b$, T)
& \checkmark & T & \checkmark & \checkmark & 
& 59.5 \gain{+7.7} & 42.7 \gain{+7.2}
& 54.4 \gain{+3.9} & 38.6 \gain{+4.9} \\


\rowcolor{cyan!15}
\textbf{REMAP}: T+O priors ($b$) + partial penalty ($\tau$)
& \checkmark & T+O & \checkmark & \checkmark & \checkmark
& \textbf{59.6} \gain{+0.5} & \textbf{42.7} \gain{+0.0}
& \textbf{59.7} \gain{+5.3} & \textbf{43.7} \gain{+5.1} \\

\hline
\bottomrule
\end{tabular}%
}
\vspace{-0.2em}
\end{table}

\begin{wraptable}{r}{0.5\columnwidth}
\vspace{-1.2em}
\centering
\caption{Sink mass is inversely correlated with the proportion of foreground content, indicating adaptive behavior based on dataset characteristics.}
\label{tab:sink_analysis}
\resizebox{0.5\columnwidth}{!}{%
\begin{tabular}{>{\rule[-0.1cm]{0pt}{0.4cm}}l|c|c}
\toprule
\textbf{Dataset} & \textbf{Foreground Ratio} & \textbf{Avg Sink Mass} \\
\midrule
PC Assembly     & \cellcolor{lightgray!35} 0.79 & \cellcolor{BurntOrange!15} 0.21 \\
MECCANO         & \cellcolor{lightgray!35} 0.50 & \cellcolor{BurntOrange!15} 0.34 \\
EGTEA-GAZE+     & \cellcolor{lightgray!35} 0.23 & \cellcolor{BurntOrange!15} 0.50 \\
\hline
\bottomrule
\end{tabular}%
}
\vspace{-0.5em}
\end{wraptable}

\vspace{-1em}
\paragraph{Analysis of Virtual Sink Behavior.}
Table~\ref{tab:sink_analysis} reports the average transport mass assigned to the virtual sink across datasets with different foreground ratios. Sink mass increases as foreground ratio decreases, ranging from $\textbf{0.21}$ on PC Assembly ($0.76$ FG) to $\textbf{0.50}$ on EGTEA-GAZE+($0.23$ FG). This suggests that the sink behaves as a \textit{data-adaptive slack variable}, selectively absorbing unmatched or low-confidence frames while preserving structured correspondences for key procedural steps. Qualitatively, sink assignments are concentrated around ambiguous, transitional, or redundant segments such as tool switching or idle motion, thereby avoiding forced correspondences between unrelated frames.

\vspace{-0.6em}
\paragraph{Effect of Clustering Methods.}
We evaluate the impact of the post-hoc segmentation stage by replacing graph-cut with K-Means, subset selection (SS), and random assignment. Graph-cut consistently produces cleaner temporal segments because it combines embedding similarity with temporal smoothness constraints. Importantly, this stage is applied only after OT optimization and does not affect the learned transport objective itself. Therefore, Table~\ref{tab:table_cluster} should be interpreted as evaluating the decoder used to convert learned embeddings into discrete key-step labels rather than explaining the alignment gains. Moreover, REMAP consistently improves over RGWOT under the same clustering strategy, indicating the \textcolor{ForestGreen}{observed gains} from the proposed transport formulation.


\begin{table}[!ht]
\centering
\vspace{-1em}
\caption{Analysis of clustering algorithms across datasets.}
\vspace{0.2em}
\label{tab:table_cluster}
\resizebox{\linewidth}{!}{%
\begin{tabular}{>{\rule[-0.1cm]{0pt}{0.4cm}}lcccccccccccccc}
\toprule
 & \multicolumn{2}{c}{CMU-MMAC} &  & \multicolumn{2}{c}{EGTEA-GAZE+} &  & \multicolumn{2}{c}{MECCANO} &  & \multicolumn{2}{c}{EPIC-Tents} &  & \multicolumn{2}{c}{ProceL} \\ 
\cline{2-3} \cline{5-6} \cline{8-9} \cline{11-12} \cline{14-15}
 & F1 & IoU &  & F1 & IoU &  & F1 & IoU &  & F1 & IoU &  & F1 & IoU \\ 
\hline

Random 
& 16.0 & 7.1 
&  & 15.6 & 6.9 
&  & 13.8 & 6.4 
&  & 14.4 & 6.8 
&  & 15.5 & 7.4 \\

RGWOT + K-means 
& 32.8 & 20.0 
&  & 30.3 & 18.2 
&  & 29.5 & 17.4 
&  & 23.8 & 13.5 
&  & - & - \\

\rowcolor{cyan!5}
REMAP + K-means 
& 40.5 \gain{+7.7} & 24.3 \gain{+4.3} 
&  & 34.8 \gain{+4.5} & 23.8 \gain{+5.6} 
&  & 35.1 \gain{+5.6} & 22.9 \gain{+5.5} 
&  & 27.9 \gain{+4.1} & 16.9 \gain{+3.4} 
&  & 36.4 & 23.2 \\
\hline

RGWOT + SS 
& 41.3 & 25.3 
&  & 34.1 & 20.7 
&  & 36.3 & 22.2 
&  & 26.4 & 15.2 
&  & - & - \\

\rowcolor{cyan!5}
REMAP + SS 
& 48.2 \gain{+6.9} & 34.9 \gain{+9.6} 
&  & 51.5 \gain{+17.4} & 40.8 \gain{+20.1} 
&  & 48.7 \gain{+12.4} & 33.5 \gain{+11.3} 
&  & 31.8 \gain{+5.4} & 20.3 \gain{+5.1} 
&  & 47.1 & 33.9 \\
\hline

RGWOT + GraphCut 
& 54.4 & 38.6 
&  & 37.4 & 22.9 
&  & 59.5 & 42.7 
&  & 39.7 & 24.9 
&  & 44.3 & 29.4 \\

\rowcolor{cyan!15} 
\textbf{\textit{REMAP}} + GraphCut 
& \textbf{59.7} \gain{+5.3} & \textbf{43.7} \gain{+5.1} 
&  & \textbf{64.2} \gain{+26.8} & \textbf{49.3} \gain{+26.4} 
&  & \textbf{59.6} \gain{+0.1} & \textbf{42.7} \gain{+0.0} 
&  & \textbf{39.8} \gain{+0.1} & \textbf{25.0} \gain{+0.1} 
&  & \textbf{57.6} \gain{+13.3} & \textbf{42.6} \gain{+13.2} \\

\hline
\bottomrule
\end{tabular}
}
\vspace{-1.6em}
\end{table}

\begin{wraptable}{r}{0.4\columnwidth}
\centering
\caption{Results for key-steps $k$.}
\label{tab:table_k}
\vspace{-0.4em}
\resizebox{0.4\columnwidth}{!}{%
\begin{tabular}{>{\rule[-0.1cm]{0pt}{0.4cm}}cccccccccc}
\toprule
\multirow{2}{*}{\textit{k}} &  & \multicolumn{3}{c}{PC Assembly} &  & \multicolumn{3}{c}{PC Disassembly} \\ \cline{3-5} \cline{7-9} 
 &  & R & F1 & IoU &  & R & F1 & IoU \\ \hline
\rowcolor{cyan!10} \textbf{7}  &  & \textbf{45.1} & \textbf{41.4} & \textbf{26.3} &  & \textbf{47.2} & \textbf{42.5} & \textbf{28.6} \\
10 &  & 40.8 & 37.9 & 23.7 &  & 44.1 & 38.4 & 25.1 \\
12 &  & 38.5 & 36.8 & 22.9 &  & 42.2 & 37.3 & 24.2 \\
15 &  & 36.2 & 35.6 & 21.3 &  & 40.1 & 36.7 & 22.7 \\ \hline \bottomrule
\end{tabular}%
}
\vspace{-1em}
\end{wraptable}

\vspace{-0.8em}
\paragraph{Number of key-steps.}
Table~\ref{tab:table_k} studies the effect of the number of discovered clusters on PC Assembly and PC Disassembly. Performance remains relatively stable across a broad range of values, with $k=7$ giving the best trade-off between capturing major procedural phases and avoiding over-segmentation. Larger values ($k \geq 10$) tend to split semantically similar actions into fragmented clusters, reducing F1 and IoU. Since $k$ is used only during post-hoc segmentation, varying it does not alter the learned OT alignment. Additional sensitivity analyses are provided in Appendix~\ref{app8.2}.

\textbf{Impact of Training Data Quantity.}
Fig.~\ref{fig:4}(b) shows performance on MECCANO as a function of training data size. REMAP consistently outperforms prior methods across all regimes and achieves strong performance even with only 2-5 videos per task. Performance improves steadily as more data becomes available, reaching \textbf{59.7} F1 at 17 videos, while competing methods scale more slowly and remain consistently behind. These results highlight the data efficiency and robustness of REMAP.

\begin{wraptable}{r}{0.6\columnwidth}
\centering
\vspace{-1em}
\caption{Comparison with SOTA unsupervised AS methods. Note `-' denotes that the authors have not provided any data on those metrics.} 
\vspace{-0.2em}
\label{tab:table4}
\resizebox{0.59\columnwidth}{!}{%
\begin{tabular}{>{\rule[-0.1cm]{0pt}{0.4cm}}lccccccc}
\toprule
\multicolumn{1}{c}{Action Segmentation} & \multicolumn{3}{c}{ProceL~\cite{elhamifar2020self_procel}} &  & \multicolumn{3}{c}{CrossTask~\cite{zhukov2019crosstask}} \\ \cline{2-4} \cline{6-8} 
\multicolumn{1}{c}{(AS) benchmark} & P & R & F1 &  & P & R & F1 \\ 
\midrule
Elhamifar \& Naing \cite{elhamifar2019unsupervised} & - & - & 29.8 &  & - & - & - \\
Elhamifar \& Huynh \cite{elhamifar2020self_procel} & 9.5 & 26.7 & 14.0 &  & 10.1 & 41.6 & 16.3 \\
Fried et al.\cite{fried2020learning} & - & - & - &  & - & 28.8 & - \\
Shen et al.\cite{shen2021learning} & 16.5 & 31.8 & 21.1 &  & 15.2 &  35.5 & 21.0 \\
Dvornik et al. \cite{dvornik2022flow} & - & - & - &  & - & - & 25.3 \\
StepFormer \cite{dvornik2023stepformer} & 18.3	& 28.1 & 21.9 &  & 22.1 & 42 & 28.3 \\
\rowcolor{lightgray!30} ASOT \cite{xu2024temporally} & - & - & 33.6 &  & - & - & 34.9 \\
\rowcolor{lightgray!30} OPEL \cite{chowdhury2024opel} & 33.6 & 36.3 & 34.9 &  & 35.6 & 34.8 & 35.1 \\ 
\rowcolor{lightgray!30} RGWOT \cite{mahmood2025procedure} & 42.2 & 46.7 & 44.3 &  & 40.4 & 40.7 & 40.4 \\
\rowcolor{cyan!10} \textit{\textbf{REMAP}} (R-FGWOT) & \underline{53.5} & \underline{60.4} & \underline{56.7} &  & \underline{60.2} & \underline{61.2} & \underline{60.6} \\
\rowcolor{cyan!10} \textit{\textbf{REMAP}} (R-FPGWOT) & \textbf{54.4} & \textbf{61.5} & \textbf{57.6} &  & \textbf{60.9} & \textbf{61.9} & \textbf{61.4} \\ \hline 
\bottomrule
\end{tabular}%
}
\vspace{-4mm}
\end{wraptable}

\vspace{-0.6em}
\paragraph{Comparison with Action Segmentation Methods.}
Although related, procedure learning (PL) and action segmentation (AS) solve different problems: PL discovers a shared set of $K$ procedural steps across videos, whereas AS segments each video independently without leveraging cross-video structure. Table~\ref{tab:table4} compares REMAP against unsupervised AS methods~\cite{xu2024temporally,dvornik2023stepformer} and OT-based approaches~\cite{chowdhury2024opel,mahmood2025procedure}. REMAP achieves the \textbf{best} \textbf{precision}, \textbf{recall}, and \textbf{F1} on both \textit{ProceL} and \textit{CrossTask}, substantially outperforming prior approaches. The balanced gains across precision and recall further indicate that REMAP avoids degenerate transport solutions while preserving coherent procedural structure.

\vspace{-0.6em}
\paragraph{Additional analyses and results.}
The appendix provides supplementary implementation details, including hyperparameter settings (App.~\ref{app2}), sensitivity analyses for loss weights and cluster count $k$ (App.~\ref{app8.2}), runtime comparisons (App.~\ref{app8.5}), graph-cut energy formulation (App.~\ref{app8.1}), and comprehensive subtask-level results for EgoProceL, ProceL, and CrossTask (App.~\ref{sec:app_datasets}).

\vspace{-0.8em}
\section{Conclusion}
\vspace{-0.6em}
We introduced \textbf{REMAP}, an unsupervised procedure learning framework that formulates cross-video alignment as partial, structure-aware optimal transport. Unlike prior OT-based methods that enforce balanced or monotonic correspondences, REMAP leverages R-FPGWOT to allow unmatched frames, enabling selective alignment of semantically meaningful content while absorbing background and redundancy. Combined with temporal Laplace priors, IDM-based structural regularization, and contrastive stabilization, REMAP yields stable and interpretable transport plans that preserve procedural order without heuristic preprocessing or strict synchronization. Empirically, REMAP consistently outperforms prior work, achieving up to \textbf{11.6\% (+4.45pp)} F1 and \textbf{19.6\% (+4.73pp)} IoU gains on EgoProceL, and an average \textbf{41\% (+17.15pp)} F1 improvement on ProceL and CrossTask. These results demonstrate that partial alignment is not merely a refinement, but a necessary modeling principle for procedure learning under real-world variability.

\bibliographystyle{unsrt}
\bibliography{references}

@inproceedings{bansal2022egoprocel_pcass,
  title={My view is the best view: Procedure learning from egocentric videos},
  author={Bansal, Siddhant and Arora, Chetan and Jawahar, CV},
  booktitle={European Conference on Computer Vision},
  pages={657--675},
  year={2022},
  organization={Springer}
}

@inproceedings{bansal2024united,
  title={United We Stand, Divided We Fall: UnityGraph for Unsupervised Procedure Learning from Videos},
  author={Bansal, Siddhant and Arora, Chetan and Jawahar, CV},
  booktitle={Proceedings of the IEEE/CVF Winter Conference on Applications of Computer Vision},
  pages={6509--6519},
  year={2024}
}

@inproceedings{elhamifar2020self_procel,
  title={Self-supervised multi-task procedure learning from instructional videos},
  author={Elhamifar, Ehsan and Huynh, Dat},
  booktitle={Computer Vision--ECCV 2020: 16th European Conference, Glasgow, UK, August 23--28, 2020, Proceedings, Part XVII 16},
  pages={557--573},
  year={2020},
  organization={Springer}
}

@inproceedings{elhamifar2019unsupervised,
  title={Unsupervised procedure learning via joint dynamic summarization},
  author={Elhamifar, Ehsan and Naing, Zwe},
  booktitle={Proceedings of the IEEE/CVF International Conference on Computer Vision},
  pages={6341--6350},
  year={2019}
}

@inproceedings{alayrac2016unsupervised,
  title={Unsupervised learning from narrated instruction videos},
  author={Alayrac, Jean-Baptiste and Bojanowski, Piotr and Agrawal, Nishant and Sivic, Josef and Laptev, Ivan and Lacoste-Julien, Simon},
  booktitle={Proceedings of the IEEE conference on computer vision and pattern recognition},
  pages={4575--4583},
  year={2016}
}

@inproceedings{kukleva2019unsupervised,
  title={Unsupervised learning of action classes with continuous temporal embedding},
  author={Kukleva, Anna and Kuehne, Hilde and Sener, Fadime and Gall, Jurgen},
  booktitle={Proceedings of the IEEE/CVF Conference on Computer Vision and Pattern Recognition},
  pages={12066--12074},
  year={2019}
}

@inproceedings{carreira2017quo,
  title={Quo vadis, action recognition? A new model and the kinetics dataset},
  author={Carreira, Joao and Zisserman, Andrew},
  booktitle={proceedings of the IEEE Conference on Computer Vision and Pattern Recognition},
  pages={6299--6308},
  year={2017}
}

@inproceedings{piergiovanni2017learning,
  title={Learning latent subevents in activity videos using temporal attention filters},
  author={Piergiovanni, A and Fan, Chenyou and Ryoo, Michael},
  booktitle={Proceedings of the AAAI conference on artificial intelligence},
  volume={31},
  number={1},
  year={2017}
}

@article{simonyan2014two,
  title={Two-stream convolutional networks for action recognition in videos},
  author={Simonyan, Karen and Zisserman, Andrew},
  journal={Advances in neural information processing systems},
  volume={27},
  year={2014}
}

@inproceedings{kumar2022unsupervised,
  title={Unsupervised action segmentation by joint representation learning and online clustering},
  author={Kumar, Sateesh and Haresh, Sanjay and Ahmed, Awais and Konin, Andrey and Zia, M Zeeshan and Tran, Quoc-Huy},
  booktitle={Proceedings of the IEEE/CVF Conference on Computer Vision and Pattern Recognition},
  pages={20174--20185},
  year={2022}
}

@inproceedings{naing2020procedure,
  title={Procedure completion by learning from partial summaries},
  author={Naing, Zwe and Elhamifar, Ehsan},
  booktitle={British Machine Vision Conference},
  year={2020}
}

@inproceedings{zhou2018towards,
  title={Towards automatic learning of procedures from web instructional videos},
  author={Zhou, Luowei and Xu, Chenliang and Corso, Jason},
  booktitle={Proceedings of the AAAI Conference on Artificial Intelligence},
  volume={32},
  number={1},
  year={2018}
}

@inproceedings{zhukov2019crosstask,
  title={Cross-task weakly supervised learning from instructional videos},
  author={Zhukov, Dimitri and Alayrac, Jean-Baptiste and Cinbis, Ramazan Gokberk and Fouhey, David and Laptev, Ivan and Sivic, Josef},
  booktitle={Proceedings of the IEEE/CVF Conference on Computer Vision and Pattern Recognition},
  pages={3537--3545},
  year={2019}
}

@inproceedings{li2020set,
  title={Set-constrained viterbi for set-supervised action segmentation},
  author={Li, Jun and Todorovic, Sinisa},
  booktitle={Proceedings of the IEEE/CVF Conference on Computer Vision and Pattern Recognition},
  pages={10820--10829},
  year={2020}
}

@inproceedings{richard2018neuralnetwork,
  title={NeuralNetwork-Viterbi: A framework for weakly supervised video learning},
  author={Richard, Alexander and Kuehne, Hilde and Iqbal, Ahsan and Gall, Juergen},
  booktitle={Proceedings of the IEEE conference on Computer Vision and Pattern Recognition},
  pages={7386--7395},
  year={2018}
}

@inproceedings{chang2019d3tw,
  title={D3tw: Discriminative differentiable dynamic time warping for weakly supervised action alignment and segmentation},
  author={Chang, Chien-Yi and Huang, De-An and Sui, Yanan and Fei-Fei, Li and Niebles, Juan Carlos},
  booktitle={Proceedings of the IEEE/CVF Conference on Computer Vision and Pattern Recognition},
  pages={3546--3555},
  year={2019}
}

@inproceedings{dwibedi2019temporal,
  title={Temporal cycle-consistency learning},
  author={Dwibedi, Debidatta and Aytar, Yusuf and Tompson, Jonathan and Sermanet, Pierre and Zisserman, Andrew},
  booktitle={Proceedings of the IEEE/CVF conference on computer vision and pattern recognition},
  pages={1801--1810},
  year={2019}
}

@inproceedings{hadji2021representation,
  title={Representation learning via global temporal alignment and cycle-consistency},
  author={Hadji, Isma and Derpanis, Konstantinos G and Jepson, Allan D},
  booktitle={Proceedings of the IEEE/CVF Conference on Computer Vision and Pattern Recognition},
  pages={11068--11077},
  year={2021}
}

@inproceedings{shen2021learning,
  title={Learning to segment actions from visual and language instructions via differentiable weak sequence alignment},
  author={Shen, Yuhan and Wang, Lu and Elhamifar, Ehsan},
  booktitle={Proceedings of the IEEE/CVF Conference on Computer Vision and Pattern Recognition},
  pages={10156--10165},
  year={2021}
}

@article{chowdhury2024opel,
  title={OPEL: Optimal transport guided procedure learning},
  author={Chowdhury, Sayeed Shafayet and Chandra, Soumyadeep and Roy, Kaushik},
  journal={Advances in Neural Information Processing Systems},
  volume={37},
  pages={59984--60011},
  year={2024}
}

@article{thorpe2018introduction,
  title={Introduction to optimal transport},
  author={Thorpe, Matthew},
  journal={Notes of Course at University of Cambridge},
  volume={3},
  year={2018}
}

@inproceedings{xu2024temporally,
  title={Temporally consistent unbalanced optimal transport for unsupervised action segmentation},
  author={Xu, Ming and Gould, Stephen},
  booktitle={Proceedings of the IEEE/CVF Conference on Computer Vision and Pattern Recognition},
  pages={14618--14627},
  year={2024}
}

@article{ali2025joint,
  title={Joint Self-Supervised Video Alignment and Action Segmentation},
  author={Ali, Ali Shah and Mahmood, Syed Ahmed and Saeed, Mubin and Konin, Andrey and Zia, M Zeeshan and Tran, Quoc-Huy},
  journal={arXiv preprint arXiv:2503.16832},
  year={2025}
}

@article{mahmood2025procedure,
  title={Procedure Learning via Regularized Gromov-Wasserstein Optimal Transport},
  author={Mahmood, Syed Ahmed and Ali, Ali Shah and Ahmed, Umer and Fateh, Fawad Javed and Zia, M Zeeshan and Tran, Quoc-Huy},
  journal={arXiv preprint arXiv:2507.15540},
  year={2025}
}

@inproceedings{peyre2016gromov,
  title={Gromov-wasserstein averaging of kernel and distance matrices},
  author={Peyr{\'e}, Gabriel and Cuturi, Marco and Solomon, Justin},
  booktitle={International conference on machine learning},
  pages={2664--2672},
  year={2016},
  organization={PMLR}
}

@article{bai2025fused,
  title={Fused Partial Gromov-Wasserstein for Structured Objects},
  author={Bai, Yikun and Tran, Huy and Du, Hengrong and Liu, Xinran and Kolouri, Soheil},
  journal={arXiv preprint arXiv:2502.09934},
  year={2025}
}

@article{boykov2002fast,
  title={Fast approximate energy minimization via graph cuts},
  author={Boykov, Yuri and Veksler, Olga and Zabih, Ramin},
  journal={IEEE Transactions on pattern analysis and machine intelligence},
  volume={23},
  number={11},
  pages={1222--1239},
  year={2002},
  publisher={IEEE}
}

@inproceedings{larsson2016learning,
  title={Learning representations for automatic colorization},
  author={Larsson, Gustav and Maire, Michael and Shakhnarovich, Gregory},
  booktitle={European conference on computer vision},
  pages={577--593},
  year={2016},
  organization={Springer}
}

@inproceedings{huang2016connectionist,
  title={Connectionist temporal modeling for weakly supervised action labeling},
  author={Huang, De-An and Fei-Fei, Li and Niebles, Juan Carlos},
  booktitle={European conference on computer Vision},
  pages={137--153},
  year={2016},
  organization={Springer}
}

@inproceedings{carlucci2019domain,
  title={Domain generalization by solving jigsaw puzzles},
  author={Carlucci, Fabio M and D'Innocente, Antonio and Bucci, Silvia and Caputo, Barbara and Tommasi, Tatiana},
  booktitle={Proceedings of the IEEE/CVF conference on computer vision and pattern recognition},
  pages={2229--2238},
  year={2019}
}

@inproceedings{kim2018learning,
  title={Learning image representations by completing damaged jigsaw puzzles},
  author={Kim, Dahun and Cho, Donghyeon and Yoo, Donggeun and Kweon, In So},
  booktitle={2018 IEEE winter conference on applications of computer vision (WACV)},
  pages={793--802},
  year={2018},
  organization={IEEE}
}

@inproceedings{kim2019self,
  title={Self-supervised video representation learning with space-time cubic puzzles},
  author={Kim, Dahun and Cho, Donghyeon and Kweon, In So},
  booktitle={Proceedings of the AAAI conference on artificial intelligence},
  volume={33},
  number={01},
  pages={8545--8552},
  year={2019}
}

@article{gidaris2018unsupervised,
  title={Unsupervised representation learning by predicting image rotations},
  author={Gidaris, Spyros and Singh, Praveer and Komodakis, Nikos},
  journal={arXiv preprint arXiv:1803.07728},
  year={2018}
}

@inproceedings{feng2019self,
  title={Self-supervised representation learning by rotation feature decoupling},
  author={Feng, Zeyu and Xu, Chang and Tao, Dacheng},
  booktitle={Proceedings of the IEEE/CVF Conference on Computer Vision and Pattern Recognition},
  pages={10364--10374},
  year={2019}
}

@inproceedings{caron2018deep,
  title={Deep clustering for unsupervised learning of visual features},
  author={Caron, Mathilde and Bojanowski, Piotr and Joulin, Armand and Douze, Matthijs},
  booktitle={Proceedings of the European conference on computer vision (ECCV)},
  pages={132--149},
  year={2018}
}

@inproceedings{caron2019unsupervised,
  title={Unsupervised pre-training of image features on non-curated data},
  author={Caron, Mathilde and Bojanowski, Piotr and Mairal, Julien and Joulin, Armand},
  booktitle={Proceedings of the IEEE/CVF International Conference on Computer Vision},
  pages={2959--2968},
  year={2019}
}

@article{ahsan2018discrimnet,
  title={Discrimnet: Semi-supervised action recognition from videos using generative adversarial networks},
  author={Ahsan, Unaiza and Sun, Chen and Essa, Irfan},
  journal={arXiv preprint arXiv:1801.07230},
  year={2018}
}

@inproceedings{diba2019dynamonet,
  title={Dynamonet: Dynamic action and motion network},
  author={Diba, Ali and Sharma, Vivek and Gool, Luc Van and Stiefelhagen, Rainer},
  booktitle={Proceedings of the IEEE/CVF international conference on computer vision},
  pages={6192--6201},
  year={2019}
}

@inproceedings{han2019video,
  title={Video representation learning by dense predictive coding},
  author={Han, Tengda and Xie, Weidi and Zisserman, Andrew},
  booktitle={Proceedings of the IEEE/CVF international conference on computer vision workshops},
  year={2019}
}

@inproceedings{srivastava2015unsupervised,
  title={Unsupervised learning of video representations using lstms},
  author={Srivastava, Nitish and Mansimov, Elman and Salakhudinov, Ruslan},
  booktitle={International conference on machine learning},
  pages={843--852},
  year={2015},
  organization={PMLR}
}

@inproceedings{fernando2017self,
  title={Self-supervised video representation learning with odd-one-out networks},
  author={Fernando, Basura and Bilen, Hakan and Gavves, Efstratios and Gould, Stephen},
  booktitle={Proceedings of the IEEE conference on computer vision and pattern recognition},
  pages={3636--3645},
  year={2017}
}

@inproceedings{lee2017unsupervised,
  title={Unsupervised representation learning by sorting sequences},
  author={Lee, Hsin-Ying and Huang, Jia-Bin and Singh, Maneesh and Yang, Ming-Hsuan},
  booktitle={Proceedings of the IEEE international conference on computer vision},
  pages={667--676},
  year={2017}
}

@inproceedings{misra2016shuffle,
  title={Shuffle and learn: unsupervised learning using temporal order verification},
  author={Misra, Ishan and Zitnick, C Lawrence and Hebert, Martial},
  booktitle={European conference on computer vision},
  pages={527--544},
  year={2016},
  organization={Springer}
}

@inproceedings{xu2019self,
  title={Self-supervised spatiotemporal learning via video clip order prediction},
  author={Xu, Dejing and Xiao, Jun and Zhao, Zhou and Shao, Jian and Xie, Di and Zhuang, Yueting},
  booktitle={Proceedings of the IEEE/CVF conference on computer vision and pattern recognition},
  pages={10334--10343},
  year={2019}
}

@inproceedings{benaim2020speednet,
  title={Speednet: Learning the speediness in videos},
  author={Benaim, Sagie and Ephrat, Ariel and Lang, Oran and Mosseri, Inbar and Freeman, William T and Rubinstein, Michael and Irani, Michal and Dekel, Tali},
  booktitle={Proceedings of the IEEE/CVF conference on computer vision and pattern recognition},
  pages={9922--9931},
  year={2020}
}

@inproceedings{wang2020self,
  title={Self-supervised video representation learning by pace prediction},
  author={Wang, Jiangliu and Jiao, Jianbo and Liu, Yun-Hui},
  booktitle={European conference on computer vision},
  pages={504--521},
  year={2020},
  organization={Springer}
}

@inproceedings{yao2020video,
  title={Video playback rate perception for self-supervised spatio-temporal representation learning},
  author={Yao, Yuan and Liu, Chang and Luo, Dezhao and Zhou, Yu and Ye, Qixiang},
  booktitle={Proceedings of the IEEE/CVF conference on computer vision and pattern recognition},
  pages={6548--6557},
  year={2020}
}

@inproceedings{vidalmata2021joint,
  title={Joint visual-temporal embedding for unsupervised learning of actions in untrimmed sequences},
  author={Vidal Mata, Rosaura G and Scheirer, Walter J and Kukleva, Anna and Cox, David and Kuehne, Hilde},
  booktitle={Proceedings of the IEEE/CVF Winter Conference on Applications of Computer Vision},
  pages={1238--1247},
  year={2021}
}

@inproceedings{damen2014you,
  title={You-Do, I-Learn: Discovering Task Relevant Objects and their Modes of Interaction from Multi-User Egocentric Video.},
  author={Damen, Dima and Leelasawassuk, Teesid and Haines, Osian and Calway, Andrew and Mayol-Cuevas, Walterio W},
  booktitle={BMVC},
  volume={2},
  pages={3},
  year={2014}
}

@inproceedings{doughty2020action,
  title={Action modifiers: Learning from adverbs in instructional videos},
  author={Doughty, Hazel and Laptev, Ivan and Mayol-Cuevas, Walterio and Damen, Dima},
  booktitle={Proceedings of the IEEE/CVF Conference on Computer Vision and Pattern Recognition},
  pages={868--878},
  year={2020}
}

@article{fried2020learning,
  title={Learning to segment actions from observation and narration},
  author={Fried, Daniel and Alayrac, Jean-Baptiste and Blunsom, Phil and Dyer, Chris and Clark, Stephen and Nematzadeh, Aida},
  journal={arXiv preprint arXiv:2005.03684},
  year={2020}
}

@article{malmaud2015s,
  title={What's cookin'? interpreting cooking videos using text, speech and vision},
  author={Malmaud, Jonathan and Huang, Jonathan and Rathod, Vivek and Johnston, Nick and Rabinovich, Andrew and Murphy, Kevin},
  journal={arXiv preprint arXiv:1503.01558},
  year={2015}
}

@inproceedings{yu2014instructional,
  title={Instructional videos for unsupervised harvesting and learning of action examples},
  author={Yu, Shoou-I and Jiang, Lu and Hauptmann, Alexander},
  booktitle={Proceedings of the 22nd ACM international conference on Multimedia},
  pages={825--828},
  year={2014}
}

@inproceedings{shah2023steps,
  title={STEPS: Self-supervised key step extraction and localization from unlabeled procedural videos},
  author={Shah, Anshul and Lundell, Benjamin and Sawhney, Harpreet and Chellappa, Rama},
  booktitle={Proceedings of the IEEE/CVF International Conference on Computer Vision},
  pages={10375--10387},
  year={2023}
}

@inproceedings{andrew2013deep,
  title={Deep canonical correlation analysis},
  author={Andrew, Galen and Arora, Raman and Bilmes, Jeff and Livescu, Karen},
  booktitle={International conference on machine learning},
  pages={1247--1255},
  year={2013},
  organization={PMLR}
}

@inproceedings{haresh2021learning,
  title={Learning by aligning videos in time},
  author={Haresh, Sanjay and Kumar, Sateesh and Coskun, Huseyin and Syed, Shahram N and Konin, Andrey and Zia, Zeeshan and Tran, Quoc-Huy},
  booktitle={Proceedings of the IEEE/CVF Conference on Computer Vision and Pattern Recognition},
  pages={5548--5558},
  year={2021}
}

@inproceedings{donahue2024learning,
  title={Learning to predict activity progress by self-supervised video alignment},
  author={Donahue, Gerard and Elhamifar, Ehsan},
  booktitle={Proceedings of the IEEE/CVF Conference on Computer Vision and Pattern Recognition},
  pages={18667--18677},
  year={2024}
}

@inproceedings{liu2022learning,
  title={Learning to align sequential actions in the wild},
  author={Liu, Weizhe and Tekin, Bugra and Coskun, Huseyin and Vineet, Vibhav and Fua, Pascal and Pollefeys, Marc},
  booktitle={Proceedings of the IEEE/CVF Conference on Computer Vision and Pattern Recognition},
  pages={2181--2191},
  year={2022}
}

@article{cuturi2013sinkhorn,
  title={Sinkhorn distances: Lightspeed computation of optimal transport},
  author={Cuturi, Marco},
  journal={Advances in neural information processing systems},
  volume={26},
  year={2013}
}

@book{villani2009optimal,
  title={Optimal transport: old and new},
  author={Villani, C{\'e}dric and others},
  volume={338},
  year={2009},
  publisher={Springer}
}

@article{albregtsen2008statistical,
  title={Statistical texture measures computed from gray level coocurrence matrices},
  author={Albregtsen, Fritz and others},
  journal={Image processing laboratory, department of informatics, university of oslo},
  volume={5},
  number={5},
  year={2008}
}

@article{greig1989exact,
  title={Exact maximum a posteriori estimation for binary images},
  author={Greig, Dorothy M and Porteous, Bruce T and Seheult, Allan H},
  journal={Journal of the Royal Statistical Society Series B: Statistical Methodology},
  volume={51},
  number={2},
  pages={271--279},
  year={1989},
  publisher={Oxford University Press}
}

@article{kuhn1955hungarian,
  title={The Hungarian method for the assignment problem},
  author={Kuhn, Harold W},
  journal={Naval research logistics quarterly},
  volume={2},
  number={1-2},
  pages={83--97},
  year={1955},
  publisher={Wiley Online Library}
}

@inproceedings{dvornik2022flow,
  title={Flow graph to video grounding for weakly-supervised multi-step localization},
  author={Dvornik, Nikita and Hadji, Isma and Pham, Hai and Bhatt, Dhaivat and Martinez, Brais and Fazly, Afsaneh and Jepson, Allan D},
  booktitle={European Conference on Computer Vision},
  pages={319--335},
  year={2022},
  organization={Springer}
}

@inproceedings{dvornik2023stepformer,
  title={StepFormer: Self-supervised step discovery and localization in instructional videos},
  author={Dvornik, Nikita and Hadji, Isma and Zhang, Ran and Derpanis, Konstantinos G and Wildes, Richard P and Jepson, Allan D},
  booktitle={Proceedings of the IEEE/CVF Conference on Computer Vision and Pattern Recognition},
  pages={18952--18961},
  year={2023}
}

@article{de2009guide_cmummac,
  title={Guide to the carnegie mellon university multimodal activity (cmu-mmac) database},
  author={De la Torre, Fernando and Hodgins, Jessica and Bargteil, Adam and Martin, Xavier and Macey, Justin and Collado, Alex and Beltran, Pep},
  year={2009},
  publisher={Citeseer}
}

@inproceedings{ragusa2021meccano,
  title={The meccano dataset: Understanding human-object interactions from egocentric videos in an industrial-like domain},
  author={Ragusa, Francesco and Furnari, Antonino and Livatino, Salvatore and Farinella, Giovanni Maria},
  booktitle={Proceedings of the IEEE/CVF Winter Conference on Applications of Computer Vision},
  pages={1569--1578},
  year={2021}
}

@inproceedings{li2018eye_egteagaze,
  title={In the eye of beholder: Joint learning of gaze and actions in first person video},
  author={Li, Yin and Liu, Miao and Rehg, James M},
  booktitle={Proceedings of the European conference on computer vision (ECCV)},
  pages={619--635},
  year={2018}
}

@inproceedings{jang2019epictents,
  title={Epic-tent: An egocentric video dataset for camping tent assembly},
  author={Jang, Youngkyoon and Sullivan, Brian and Ludwig, Casimir and Gilchrist, Iain and Damen, Dima and Mayol-Cuevas, Walterio},
  booktitle={Proceedings of the IEEE/CVF International Conference on Computer Vision Workshops},
  year={2019}
}

@article{adam2014method,
  title={Adam: A method for stochastic optimization},
  author={Kingma, Diederik P and Ba, Jimmy},
  journal={arXiv preprint arXiv:1412.6980},
  year={2014}
}

@article{Sinkhorn1967,
  author    = {R. Sinkhorn},
  title     = {Diagonal equivalence to matrices with prescribed row and column sums},
  journal   = {The American Mathematical Monthly},
  year      = {1967},
  volume    = {74},
  number    = {4},
  pages     = {402--405}
}

@article{BorobiaCanto1998,
  author    = {A. Borobia and R. Cant{\'o}},
  title     = {Matrix scaling: A geometric proof of Sinkhorn's theorem},
  journal   = {Linear Algebra and its Applications},
  year      = {1998},
  volume    = {268},
  pages     = {1--8}
}

@article{chizat2018scaling,
  title={Scaling algorithms for unbalanced optimal transport problems},
  author={Chizat, Lenaic and Peyr{\'e}, Gabriel and Schmitzer, Bernhard and Vialard, Fran{\c{c}}ois-Xavier},
  journal={Mathematics of computation},
  volume={87},
  number={314},
  pages={2563--2609},
  year={2018}
}



\appendix


\section{Appendix}
\subsection{Derivation and Optimization of R-FPGWOT $(\hat{\boldsymbol{T}}_{\lambda_1,\lambda_2})$}
\label{app:proof_fpgwot}

This appendix derives the optimization scheme for \textit{Regularized Fused Partial Gromov-Wasserstein Optimal Transport} (R-FPGWOT). We first define the augmented objective, then derive the MM linearization, the resulting unbalanced Sinkhorn updates, and finally discuss convergence, complexity, and temporal subsampling. We also clarify how the virtual sink is parameterized and how temporal structure matrices are padded to avoid sink-position bias.

\vspace{-0.3em}
\paragraph{Notation.}
Let $X=\{x_i\}_{i=1}^N$ and $Y=\{y_j\}_{j=1}^M$ denote frame embeddings from two videos, stacked as $\boldsymbol X\in\mathbb{R}^{N\times d}$ and $\boldsymbol Y\in\mathbb{R}^{M\times d}$. We optimize an augmented coupling
$\hat{\boldsymbol T}\in\mathbb{R}_+^{(N+1)\times(M+1)}$, where the additional row and column correspond to a virtual sink for unmatched mass. Let $\boldsymbol 1_k$ denote a vector of ones.

\vspace{-0.3em}
\paragraph{Costs and virtual sink.}
The real-frame appearance cost is $\boldsymbol C_{ij}=\|x_i-y_j\|_2$. We augment it as:
\begin{equation}
\hat{\boldsymbol C}
=
\begin{bmatrix}
\boldsymbol C & \zeta\boldsymbol 1_N\\
\zeta\boldsymbol 1_M^\top & 0
\end{bmatrix},
\label{eq:app_sink_cost}
\notag
\end{equation}
where $\zeta$ is the finite virtual-frame threshold. A smaller $\zeta$ makes sink assignment easier, while a larger $\zeta$ encourages real-frame matching unless the real match is unreliable.

The temporal structure matrices are padded as:
\begin{equation}
\hat{\boldsymbol C}^x=
\begin{bmatrix}
\boldsymbol C^x & \boldsymbol 0\\
\boldsymbol 0^\top & 0
\end{bmatrix},
\qquad
\hat{\boldsymbol C}^y=
\begin{bmatrix}
\boldsymbol C^y & \boldsymbol 0\\
\boldsymbol 0^\top & 0
\end{bmatrix}.
\label{eq:app_sink_structure}
\notag
\end{equation}
Thus, the virtual sink is not treated as the last chronological frame. It is masked from temporal similarity computation and remains position-invariant.

\vspace{-0.3em}
\paragraph{Priors and marginals.}
We use a strictly positive augmented prior $\hat{\boldsymbol Q}\in\mathbb{R}_{++}^{(N+1)\times(M+1)}$ and positive target marginals $\hat{\boldsymbol\alpha}\in\Delta^{N+1}$ and $\hat{\boldsymbol\beta}\in\Delta^{M+1}$, including virtual mass. For real-frame entries, $\hat Q_{ij}=Q_{ij}$ for $i\le N,j\le M$. For sink entries, we set:
\[
\hat Q_{i,M+1}=q_{\mathrm{sink}},\qquad
\hat Q_{N+1,j}=q_{\mathrm{sink}},\qquad
\hat Q_{N+1,M+1}=q_{\mathrm{ss}},
\]
with small positive constants $q_{\mathrm{sink}},q_{\mathrm{ss}}>0$. 

We define generalized KL divergence as:
\[
\mathrm{KL}(\boldsymbol A\Vert\boldsymbol B)
=
\sum_{ij} A_{ij}\log\frac{A_{ij}}{B_{ij}}-A_{ij}+B_{ij}.
\]

\subsubsection*{I. \; \underline{R-FPGWOT Objective}}
\label{app:objective}

The constrained form of R-FPGWOT combines appearance matching with a structural GW reward. Since $\hat{\boldsymbol C}^x$ and $\hat{\boldsymbol C}^y$ are temporal similarity kernels, structurally consistent matches should be rewarded, not penalized. Therefore, the GW term appears with a negative sign inside the minimization objective:
\begin{equation}
\label{eq:app_constrained}
\begin{aligned}
l^{R-FPGW}_{\xi_1,\xi_2}(\boldsymbol X,\boldsymbol Y)
\;=\;
\min_{\hat{\boldsymbol T}\ge 0}\;&
(1-\rho)\,\underbrace{\langle \boldsymbol C,\hat{\boldsymbol T}\rangle}_{\text{KOT term}}
\ +\ 
\rho\,\underbrace{\langle \boldsymbol C^x \hat{\boldsymbol T}\boldsymbol C^y,\hat{\boldsymbol T}\rangle}_{\text{GW term}}
\;+\;\underbrace{\tau\Big[\mathrm{KL}(\hat{\boldsymbol T}\boldsymbol 1\Vert\alpha)+\mathrm{KL}(\hat{\boldsymbol T}^\top\boldsymbol 1\Vert\beta)\Big]}_{\text{Marginal Penalties}} \\
\text{s.t. }&
M(\hat{\boldsymbol T}) \;\ge\; \xi_1,
\qquad
\mathrm{KL}(\hat{\boldsymbol T}\Vert\hat{\boldsymbol Q}) \;\le\; \xi_2 .
\end{aligned}
\tag{A1}
\end{equation}

Taking the Lagrangian of Eq.~\ref{eq:app_constrained} introduces multipliers $\lambda_1,\lambda_2>0$, yielding the penalized objective:
\begin{equation}
\label{eq:app:master}
\begin{aligned}
\mathcal J(\hat{\boldsymbol T})
=
(1-\rho)\langle &\boldsymbol C,\hat{\boldsymbol T}\rangle
+\rho\langle \boldsymbol C^x\hat{\boldsymbol T}\boldsymbol C^y,\hat{\boldsymbol T}\rangle
-\lambda_1 M(\hat{\boldsymbol T})
+\lambda_2\mathrm{KL}(\hat{\boldsymbol T}\Vert\hat{\boldsymbol Q}) \\
&+\tau\Big[
\mathrm{KL}(\hat{\boldsymbol T}\boldsymbol 1_{M+1}\Vert\boldsymbol\alpha)
+\mathrm{KL}(\hat{\boldsymbol T}^{\top}\boldsymbol 1_{N+1}\Vert\boldsymbol\beta)
\Big].
\end{aligned}
\tag{A2}
\end{equation}

Here, $M(\hat{\boldsymbol T})$ is the IDM-style structural reward. Sink entries are excluded from $M(\hat{\boldsymbol T})$, so the sink does not receive artificial temporal preference.

\subsubsection*{II. \; \underline{MM Linearization of the Fused Term}}
\label{app:mm}
Let 
$F(\boldsymbol T)=\langle \boldsymbol C^x \boldsymbol T \boldsymbol C^y,\boldsymbol T\rangle$
denote the fused GW quadratic term. At outer iteration $s$, we linearize this term around
the current transport plan $\hat{\boldsymbol T}^{(s)}$. Since $F$ has Lipschitz-continuous
gradient, it admits the standard first-order quadratic majorizer
\begin{equation}
F(\boldsymbol T)
\le
F(\hat{\boldsymbol T}^{(s)})
+
\left\langle
\boldsymbol G^{(s)},\boldsymbol T-\hat{\boldsymbol T}^{(s)}
\right\rangle
+
\frac{L}{2}
\left\|\boldsymbol T-\hat{\boldsymbol T}^{(s)}\right\|_F^2,
\qquad
\boldsymbol G^{(s)}=\nabla F(\hat{\boldsymbol T}^{(s)}),
\label{eq:app:majorizer}
\tag{A3}
\end{equation}
where $L$ is a Lipschitz constant of $\nabla F$. 

For symmetric structure kernels ($\boldsymbol C^x$ and $\boldsymbol C^y$), the gradient is:
\begin{equation}
\boldsymbol G^{(s)}
=
2\hat{\boldsymbol C}^x\hat{\boldsymbol T}^{(s)}\hat{\boldsymbol C}^y .
\label{eq:app:gw_gradient_sym}
\tag{A4}
\end{equation}

More generally, for arbitrary bounded structure matrices,
\begin{equation}
\boldsymbol G^{(s)}
=
\hat{\boldsymbol C}^x\hat{\boldsymbol T}^{(s)}(\hat{\boldsymbol C}^y)^\top
+
(\hat{\boldsymbol C}^x)^\top\hat{\boldsymbol T}^{(s)}\hat{\boldsymbol C}^y .
\label{eq:app:gw_gradient_general}
\tag{A5}
\end{equation}

Therefore, the effective linearized cost used in the inner OT problem is:
\begin{equation}
\widetilde{\boldsymbol D}^{(s)}
=
(1-\rho)\hat{\boldsymbol C}
-
\rho\boldsymbol G^{(s)} .
\label{eq:app:linearized_cost}
\tag{A6}
\end{equation}
The negative sign is important: high structural similarity lowers the effective transport cost.
Thus, the update can be interpreted as an MM step: with PSD temporal kernels it yields a global quadratic majorizer, while for bounded non-PSD temporal distances it yields a local surrogate with standard stationary-point guarantees.

\vspace{-0.3em}
\paragraph{Temporal structure matrices.}
For the theoretical descent argument, we use PSD temporal similarity kernels:
\begin{equation}
(\boldsymbol C^x)_{ii'}
=
\exp\!\left(-\frac{|i-i'|}{b_x}\right),
\qquad
(\boldsymbol C^y)_{jj'}
=
\exp\!\left(-\frac{|j-j'|}{b_y}\right),
\tag{A7}
\end{equation}
or Gaussian kernels over temporal indices. These kernels are symmetric and positive semidefinite. Since $F$ is convex under PSD kernels, $-F$ is concave, and its first-order Taylor expansion is a global upper bound. Thus, replacing $-F$ by its tangent yields a valid MM majorizer for the minimization objective. If raw temporal distance matrices are used instead, they are generally not PSD; in that case, the same update can be interpreted as a local entropic FGW surrogate with stationary-point convergence under standard smoothness assumptions. In all experiments, we use the PSD-kernel construction.

\subsubsection*{III. \; \underline{Inner (Convex) Unbalanced OT Subproblem and Gibbs Kernel Formulation}}
\label{app:inner}
We start from the unconstrained KL-regularized formulation (ignoring additive constants). 
The objective combines (i) linearized cost, (ii) IDM reward, (iii) prior-KL, and (iv) marginal KL penalties (for the unbalanced case). R-FPGWOT inherits the convergence behavior of entropic fused GW methods: while global optimality is not guaranteed due to non-convexity, the alternating linearization with Sinkhorn updates converges to a stationary point under mild assumptions. Our constraints act as soft regularizers rather than hard projections, preserving feasibility and numerical stability.

\paragraph{General inner problem.} 
Fixing $\widetilde{\boldsymbol D}^{(s)}$, the inner problem at iteration $s$ is:
\begin{equation}
\min_{\hat{\boldsymbol T}\ge 0}\quad
\big\langle \hat{\boldsymbol T},\,\widetilde{\boldsymbol D}^{(s)}\big\rangle
-\lambda_1 M(\hat{\boldsymbol T})
+\lambda_2\,\mathrm{KL}(\hat{\boldsymbol T}\Vert\hat{\boldsymbol Q}^{(s)})
\nonumber
+\tau\Big(
\mathrm{KL}(\hat{\boldsymbol T}\boldsymbol 1_{M+1}\Vert\hat{\boldsymbol\alpha})
+
\mathrm{KL}(\hat{\boldsymbol T}^{\top}\boldsymbol 1_{N+1}\Vert\hat{\boldsymbol\beta})
\Big).
\label{eq:app:inner}
\tag{A8} 
\end{equation}
Since $\lambda_2>0$ and $\hat{\boldsymbol Q}^{(s)}>0$, the prior KL term is strictly convex in $\hat{\boldsymbol T}$, giving a unique minimizer for the inner subproblem.

\paragraph{IDM score.}
We write the IDM reward as a linear negative score:
\[
M(\hat{\boldsymbol T})
=
\sum_{i=1}^{N}\sum_{j=1}^{M} s_{ij}^{(s)}\hat T_{ij},
\]
where;
\[
s_{ij}^{(s)}
=\left[
\phi^{(s)}
\frac{1}{(\tfrac{i}{N+1}-\tfrac{j}{M+1})^2+1}
+
(1-\phi^{(s)})
\frac{1}{\tfrac{1}{2}d_m(i,j)+1}
\right];
\qquad 
d_m=\!\left(\tfrac{i-i_o}{N+1}\right)^2+\left(\tfrac{j-j_o}{M+1}\right)^2
\]
The sink row and column are masked by setting $S_{ij}^{(s)}=0$ whenever $i=N+1$ or $j=M+1$.

\textbf{Lagrangian.}
Let $r_i=(\hat{\boldsymbol T}\boldsymbol 1)_i$ and $c_j=(\hat{\boldsymbol T}^{\top}\boldsymbol 1)_j$ denote the row and column sums. Keeping the full generalized KL terms, the inner objective is:
\begin{equation}
\label{eq:A5}
\begin{aligned}
\mathcal{L}(\hat{\boldsymbol T})
&=
\sum_{i,j}\widetilde d^{(s)}_{ij}t_{ij}
-
\lambda_1\sum_{i,j}s_{ij}^{(s)}t_{ij}
+
\lambda_2\sum_{i,j}
\left(
t_{ij}\log\frac{t_{ij}}{q_{ij}^{(s)}}-t_{ij}+q_{ij}^{(s)}
\right)
\\
&\quad+
\tau\left[
\sum_i
\left(
r_i\log\frac{r_i}{\hat\alpha_i}-r_i+\hat\alpha_i
\right)
+
\sum_j
\left(
c_j\log\frac{c_j}{\hat\beta_j}-c_j+\hat\beta_j
\right)
\right].
\end{aligned}
\tag{A9}
\end{equation}

where, $t_{ij}$ is transport $(i,j)$ entry and $q_{ij}>0$ is prior $(i,j)$ entry of $\hat{\boldsymbol T}$ and $\hat{\boldsymbol Q}$ respectively and $\lambda_2>0$ is the temperature.

\textbf{Stationarity (KKT).}  
Differentiating Eq.~\ref{eq:A5} with respect to $t_{ij}$ and setting $\partial \mathcal{L}/\partial t_{ij}=0$, yields the stationary condition:
\begin{equation}
\frac{\partial \mathcal{L}}{\partial t_{ij}}
=
\widetilde d^{(s)}_{ij}
-
\lambda_1 s_{ij}^{(s)}
+
\lambda_2\log\frac{t_{ij}}{q_{ij}^{(s)}}
+
\tau\left(
\log\frac{r_i}{\hat\alpha_i}
+
\log\frac{c_j}{\hat\beta_j}
\right)
=0 .
\label{eq:A6}
\tag{A10}
\end{equation}

\textbf{Gibbs form.}
Rearranging Eq.~\ref{eq:A6}, the KKT stationarity yields a Gibbs form:
\begin{equation}
t_{ij}
=
K^{(s)}_{ij}
\left(\frac{\hat\alpha_i}{r_i}\right)^{\eta}
\left(\frac{\hat\beta_j}{c_j}\right)^{\eta},
\qquad
\eta=\frac{\tau}{\lambda_2},
\label{eq:app:gibbs}
\tag{A11}
\end{equation}
with strictly positive Gibbs kernel,
\vspace{-0.5em}
\begin{equation}
K^{(s)}_{ij}
=
q_{ij}^{(s)}
\exp\left(
\frac{\lambda_1 s_{ij}^{(s)}-\widetilde d^{(s)}_{ij}}{\lambda_2}
\right).
\label{eq:app:kernel}
\tag{A12}
\end{equation}

\textbf{Unbalanced Sinkhorn Scaling (Partial OT).}
\label{app:unb}
Since $q_{ij}^{(s)}>0$ and the exponent in Eq.~\ref{eq:app:kernel} is finite, $\boldsymbol K^{(s)}$ has strictly positive entries. Therefore, by Sinkhorn's theorem (Theorem A\footnote{\textbf{Balanced Sinkhorn existence (classical).}
For any positive matrix $\boldsymbol A$, there exist positive diagonal scalings that match prescribed positive marginals (up to a common factor) \cite{Sinkhorn1967,BorobiaCanto1998}. In the unbalanced (KL-penalized) setting used here, the fixed-point equations Eq.~\ref{eq:app:unbalanced_sinkhorn} arise from KKT stationarity and admit unique positive solutions under bounded positive kernels; see, e.g., unbalanced OT analyses \cite{chizat2018scaling}.}) 
in entropic optimal transport, there exist positive scaling vectors $\boldsymbol u \in \mathbb{R}^{N+1}, \boldsymbol v \in \mathbb{R}^{M+1}$ such that:
\begin{equation}
\hat{\boldsymbol T}
=
\mathrm{Diag}(\boldsymbol u)\boldsymbol K^{(s)}\mathrm{Diag}(\boldsymbol v),
\qquad
r_i=u_i(\boldsymbol K^{(s)}\boldsymbol v)_i,
\qquad
c_j=v_j((\boldsymbol K^{(s)})^\top\boldsymbol u)_j .
\label{eq:a13}
\tag{A13}
\end{equation}

Substituting Eq.~\ref{eq:a13} into Eq.~\ref{eq:app:gibbs} gives the damped unbalanced Sinkhorn updates:
\begin{equation}
\boldsymbol u
\leftarrow
\left(
\frac{\hat{\boldsymbol\alpha}}{\boldsymbol K^{(s)}\boldsymbol v}
\right)^{\kappa},
\qquad
\boldsymbol v
\leftarrow
\left(
\frac{\hat{\boldsymbol\beta}}{(\boldsymbol K^{(s)})^\top\boldsymbol u}
\right)^{\kappa},
\qquad
\kappa=\frac{\tau}{\tau+\lambda_2}\in(0,1).
\label{eq:app:unbalanced_sinkhorn}
\tag{A14}
\end{equation}
As $\tau\rightarrow\infty$, we have $\kappa\rightarrow1$, and the updates recover balanced Sinkhorn scaling. For finite $\tau$, the marginal constraints are softly enforced, yielding partial transport.

\vspace{-0.3em}
\paragraph{Virtual sink.}
The last row and column of $\hat{\boldsymbol T}$ (index $N\!+\!1$/$M\!+\!1$ correspond to the virtual sink and are updated exactly as ordinary entries in Eq.~\ref{eq:app:unbalanced_sinkhorn}. However, unlike ordinary frames, the sink is masked from temporal priors and IDM scores. This avoids treating the sink as the final timestamp of the video. Positive virtual mass allocated in $\hat{\boldsymbol\alpha}$ and $\hat{\boldsymbol\beta}$, together with finite sink costs $\zeta$, allows low-confidence or unmatched frames to be absorbed without numerical instability.

\vspace{-0.8em}
\paragraph{Inner Convergence.}
Assume $\boldsymbol K^{(s)}$ has strictly positive bounded entries, i.e.,
$0<m\le K^{(s)}_{ij}\le M<\infty$, and $\hat{\boldsymbol\alpha},\hat{\boldsymbol\beta}$ have strictly positive components, including virtual mass. Then the unbalanced Sinkhorn updates in Eq.~\ref{eq:app:unbalanced_sinkhorn} converge to the unique minimizer of Eq.~\ref{eq:app:inner}, following standard results for KL-regularized unbalanced OT \cite{cuturi2013sinkhorn,chizat2018scaling}.

\subsubsection*{IV. \; \underline{Outer MM Convergence}}
\label{app:mm_conv}

We optimize the R-FPGWOT objective using a majorization-minimization (MM) scheme. At each outer iteration, the structural reward is linearized (Eq.~\ref{eq:app:majorizer}), and the resulting inner convex KL-regularized unbalanced OT subproblem is solved with unbalanced Sinkhorn updates. When $\hat{\boldsymbol C}^x,\hat{\boldsymbol C}^y$ are symmetric PSD kernels, $F(\hat{\boldsymbol T})=\langle \hat{\boldsymbol C}^x\hat{\boldsymbol T}\hat{\boldsymbol C}^y,\hat{\boldsymbol T}\rangle$ is convex. Therefore, $-F(\hat{\boldsymbol T})$ is concave, and its first-order Taylor expansion gives a global upper bound. This yields a valid MM update and monotone decrease of the objective:
\begin{equation}
\mathcal{J}(\hat{\boldsymbol T}^{(s+1)})
\le
\mathcal{J}(\hat{\boldsymbol T}^{(s)}).
\label{eq:app:outer_mm_decrease}
\tag{A15}
\end{equation}
For bounded non-PSD structure matrices, the same update can be interpreted as a local smooth surrogate, yielding convergence to a stationary point under standard entropic FGW assumptions. In both cases, the strictly convex inner problem ensures a unique solution at each step, and the overall procedure produces a sequence of stable transport plans with non-increasing objective values.

\vspace{-0.5em}
\paragraph{Computational complexity.}
For videos with $N$ and $M$ sampled frames, each inner Sinkhorn iteration requires two matrix-vector products with $\boldsymbol K^{(s)}$ and $(\boldsymbol K^{(s)})^\top$, costing $O((N+1)(M+1))$. The GW linearization adds the cost of recomputing $\boldsymbol G^{(s)}$. With banded or kernelized temporal structure matrices, this remains practical and avoids dense Kronecker constructions. As previous studies have shown that a small number of iterations is sufficient for effective convergence \cite{cuturi2013sinkhorn}, in our experiments, we use at most $25$ inner Sinkhorn iterations and $3$-$6$ outer MM steps. We stop the inner loop by relative marginal change ($\le 10^{-3}$) and the outer loop by relative objective decrease ($\le 10^{-4}$).


\subsubsection*{V. \; \underline{Temporal subsampling and scalability.}}

\begin{wraptable}{r}{0.45\columnwidth}
\renewcommand\thetable{A1}
\vspace{-1.2em}
\centering
\caption{Effect of temporal subsampling on MECCANO.}
\label{tab:subsampling}
\vspace{-0.4em}
\resizebox{0.45\columnwidth}{!}{%
\begin{tabular}{>{\rule[-0.1cm]{0pt}{0.4cm}}cccccc}
\toprule
\# \textbf{Frames} & \textbf{120} & \textbf{100} & \textbf{80} & \textbf{60} & \textbf{40} \\
\midrule
F1 (\%) & \cellcolor{cyan!15}\textbf{59.6} & 59.3 & 58.9 & 57.2 & 55.7 \\
\bottomrule
\end{tabular}%
}
\vspace{-1em}
\end{wraptable}

Scalability is a known limitation of OT-based methods because the pairwise transport plan scales as $O(NM)$. We mitigate this by temporal subsampling and video-pair parallelization. In all experiments, we sample at most $\textbf{120}$ frames per video, which preserves procedural structure while keeping memory and runtime comparable to prior OT-based methods such as RGWOT and ASOT. As shown in Table~\ref{tab:subsampling}, performance remains stable under moderate temporal subsampling, with only gradual degradation at lower frame counts. This indicates that REMAP does not rely on dense temporal resolution. However, very aggressive subsampling may affect fine-grained boundaries and merge short procedural steps. On an A40 GPU, aligning a pair of videos with approximately $100$ sampled frames takes $1.8$-$2.5$ seconds, and pairwise alignments can be parallelized across GPUs. Extremely long or highly fine-grained sequences remain challenging, and hierarchical or sparse OT formulations are promising future directions.

\subsubsection*{VI. \; \underline{Algorithmic Summary} (Practical Implementation)}
Algorithm~\ref{alg:rfpgwot} summarizes the practical optimization. At each outer iteration, we recompute the annealed prior $\hat{\boldsymbol Q}^{(s)}$, the IDM score $S^{(s)}$, and the linearized GW reward. The resulting unbalanced entropic OT subproblem is solved with Sinkhorn iterations.


\begin{algorithm}[H]
\caption{: R-FPGWOT with IDM Priors and Unbalanced Sinkhorn}
\label{alg:rfpgwot}
\begin{algorithmic}[1]
\State \textbf{Input:} augmented costs $\hat{\boldsymbol C},\hat{\boldsymbol C}^x,\hat{\boldsymbol C}^y$, marginals $\hat{\boldsymbol\alpha},\hat{\boldsymbol\beta}$, weights $\rho,\lambda_1,\lambda_2,\tau$, sink threshold $\zeta$, annealing schedule $\phi$.
\State Initialize $\hat{\boldsymbol T}^{(0)}$; set outer index $s\gets0$.
\Repeat \Comment{Outer MM}
    \State Recompute annealed prior $\hat{\boldsymbol Q}^{(s)}$ using $\phi^{(s)}$; keep sink entries constant.
    \State Compute GW gradient: 
    $
    \qquad \qquad \boldsymbol G^{(s)} \gets 2\,\boldsymbol C^x \hat{\boldsymbol T}^{(s)} \boldsymbol C^y
    $
    \Comment{Symmetric PSD}
    \State Form linearized fused cost: 
    $
    \widetilde{\boldsymbol D}^{(s)} \gets (1-\rho)\,\boldsymbol C + \rho\,\boldsymbol G^{(s)}
    $
    \State Compute IDM score $s^{\lambda_1}_{ij}$ for each entry: 
    \[
    s_{ij}^{(s)}
    \gets
    \phi^{(s)}
    \Big[(\tfrac{i}{N+1}-\tfrac{j}{M+1})^2+1\Big]^{-1}
    +
    (1-\phi^{(s)})
    \Big[\tfrac{1}{2}d_m+1\Big]^{-1}.
    \].
    \State Set $s_{ij}^{(s)}\gets0$ if $i=N+1$ or $j=M+1$.
    \State Build Gibbs kernel: \Comment{cf. Eq.~\ref{eq:app:kernel}}
    \[
    K^{(s)}_{ij}
    \gets
    q_{ij}^{(s)}
    \exp\left(
    \frac{\lambda_1 s_{ij}^{(s)}-\widetilde D_{ij}^{(s)}}{\lambda_2}
    \right).
    \]
    \State Initialize $\boldsymbol u\gets\boldsymbol 1$, $\boldsymbol v\gets\boldsymbol 1$; set $\kappa\gets \tau/(\tau+\lambda_2)$.
    \Repeat
        \State $\boldsymbol u\gets \left(\hat{\boldsymbol\alpha} ./(\boldsymbol K^{(s)}\boldsymbol v)\right)^{\kappa}$.
        \State $\boldsymbol v\gets \left(\hat{\boldsymbol\beta} ./((\boldsymbol K^{(s)})^\top\boldsymbol u)\right)^{\kappa}$.
    \Until{Inner convergence}
    \State Update:
    \[
    \hat{\boldsymbol T}^{(s+1)}
    \gets
    \mathrm{Diag}(\boldsymbol u)\boldsymbol K^{(s)}\mathrm{Diag}(\boldsymbol v).
    \]

    \State Anneal $\phi$; set $s\gets s+1$.
\Until{Outer converged}
\State \textbf{Return} $\hat{\boldsymbol T}^{(s)}$.
\end{algorithmic}
\end{algorithm}

\vspace{-0.3em}
\paragraph{Consistency and special cases.}
The proposed formulation recovers several standard OT variants and provide useful implementation checks:
(i) $\tau\to\infty$ reduces to a balanced FGWOT-style formulation,
(ii) $\rho=0$ reduces to entropic Kantorovich OT with partial marginal relaxation,
(iii) $\lambda_1=0$ removes IDM structural regularization,
(iv) removing the prior KL tether reduces the method to unbalanced entropic FPGWOT.
\subsection{Hyperparameter Settings}
\label{app2}
Table~\ref{tab:hyper-param} reports the default hyperparameters used for REMAP. Unless otherwise specified, the same configuration is used across all datasets. The main dataset-dependent parameter is the Laplace scale $b$, set to $3.0$ for MECCANO and EPIC-Tents and $2.0$ otherwise.

\paragraph{Evaluation key-step counts.}
To preserve a strictly unsupervised setting, the number of discovered key-step clusters is fixed to $K=7$ for all datasets and tasks. This choice is motivated by the observation that the average number of semantically distinct procedural stages across datasets is approximately seven, providing a compact yet expressive decomposition of instructional procedures. Importantly, the discovered clusters represent latent procedural stages rather than a forced recovery of every fine-grained annotation. Datasets such as MECCANO contain substantially more atomic annotations (e.g., 17 annotated steps), many of which correspond to visually similar or temporally adjacent sub-actions. In practice, multiple fine-grained annotations may therefore align to a shared higher-level procedural cluster. Evaluation is performed using Hungarian matching between discovered clusters and annotated steps following prior unsupervised procedure learning protocols.

\begin{table}[h!]
\renewcommand\thetable{A2}
\centering
\caption{Hyperparameter settings for REMAP.}
\vspace{0.4em}
\label{tab:hyper-param}
\begin{tabular}{lr}
\toprule
\textbf{Hyperparameter} & \textbf{Value} \\
\midrule
No. of key-steps ($k$) & 7 \\
No. of sampled frames ($N, M$) & 120 \\
Max. training epochs & 10000 \\
Batch size & 2 \\
Learning rate & $10^{-4}$ \\
Weight decay & $10^{-5}$ \\
Window size ($\delta$) & 15 \\
No. of context frames & 2 \\
Context stride & 15 \\
Embedding dimension & 128 \\
Optimizer & Adam~\cite{adam2014method}\\
\midrule
KOT--GWOT trade-off ($\rho$) & 0.5 \\
Entropy regularization ($\epsilon$) & 0.07 \\
Laplace scale ($b$) & 3.0 (MECCANO, EPIC-Tents) \\
Laplace scale ($b$) & 2.0 (all other datasets) \\
Temperature & 0.5 \\
$\lambda_1$ & $\frac{1}{N+M}$ \\
$\lambda_2$ & $\frac{0.1NM}{4.0}$ \\
Margin ($\lambda_3$) & 2.0 \\
Virtual-frame threshold ($\zeta$) & $\frac{10}{N+M}$ \\
$c_1$ & $\frac{1}{NM}$ \\
$c_2$ & 0.5 \\
$c_3$ & $10^{-4}$ \\
Maximum Sinkhorn iterations & 20 \\
\hline
\bottomrule
\end{tabular}
\end{table}

\subsection{Compute Resources for Experiments}
\label{app3}
For our experiments, appropriate computational resources were required to ensure efficient model training. All experiments are conducted on a single NVIDIA A40 GPU. With batch size $2$, REMAP requires approximately 16GB of GPU memory because optimization is performed on frame embeddings rather than raw video frames. Training time depends on the number of videos and sampled pairs; for tasks with 15-20 videos, such as PC Assembly or MECCANO, training takes approximately 12 hours under the default setting of 10,000 epochs. Pairwise alignment computations are independent and can be parallelized across GPUs.

\subsection{Detailed Statistics of Dataset}
Table \ref{tab:abs2} presents statistical analyses for each of the 16 (5+7+4) tasks in the EgoProceL dataset \cite{bansal2022egoprocel_pcass}. Here, $N$ denotes the total number of videos, while $K$ represents the number of key-steps for each task. \textit{$u_n$} indicates the number of unique key-steps, and \textit{$g_n$} denotes the number of annotated key-steps for the $n^{th}$ video. Following the methodology in \cite{elhamifar2019unsupervised}, these statistics characterize the amount of background clutter, skipped steps, and repetition present in each dataset:

\textit{Foreground Ratio:} This metric measures the proportion of the total video duration occupied by key-steps. It reflects the prevalence of background actions in a task. A higher foreground ratio (closer to 1) corresponds to fewer background actions. It is defined as:
\begin{equation}
    F = \frac {\sum_{n=1}^{N} \frac {t_{k}^{n}}{t_{v}^{n}}}{N} \tag{A16}
\end{equation}
where \( t_{k}^{n} \) and \( t_{v}^{n} \) denote the durations of key-steps and the full video for the $n^{th}$ instance, respectively.

\begin{table}[!ht]
\renewcommand\thetable{A3}
\centering
\caption{Statistics of the EgoProceL dataset across different tasks.}
\vspace{0.4em}
\label{tab:abs2}
\resizebox{0.82\columnwidth}{!}{%
\begin{tabular}{>{\rule[-0.2cm]{0pt}{0.5cm}}lccccc}
\toprule
\multirow{2}{*}{Task} & Videos & Key-steps & Foreground & Missing & Repeated \\
 & Count & Count & Ratio & Key-steps & Key-steps \\ \midrule
PC Assembly \cite{bansal2022egoprocel_pcass} & 14 & 9 & 0.79 & 0.02 & 0.65 \\
PC Disassembly \cite{bansal2022egoprocel_pcass} & 15 & 9 & 0.72 & 0.00 & 0.60 \\
MECCANO \cite{ragusa2021meccano} & 20 & 17 & 0.50 & 0.06 & 0.32 \\
Epic-Tents \cite{jang2019epictents} & 29 & 12 & 0.63 & 0.14 & 0.73 \\ \midrule
\underline{CMU-MMAC} \cite{de2009guide_cmummac} &  &  &  &  &  \\ 
\hspace{2em} Brownie & 34 & 9 & 0.44 & 0.19 & 0.26 \\
\hspace{2em} Eggs & 33 & 8 & 0.26 & 0.05 & 0.26 \\
\hspace{2em} Pepperoni Pizza & 33 & 5 & 0.53 & 0.00 & 0.26 \\
\hspace{2em} Salad & 34 & 9 & 0.32 & 0.30 & 0.14 \\
\hspace{2em} Sandwich & 31 & 4 & 0.25 & 0.03 & 0.37 \\ \midrule
\underline{EGTEAGAZE+} \cite{li2018eye_egteagaze} &  &  &  &  &  \\
\hspace{2em} Bacon and Eggs & 16 & 11 & 0.15 & 0.22 & 0.51 \\
\hspace{2em} Cheese Burger & 10 & 10 & 0.22 & 0.22 & 0.65 \\
\hspace{2em} Continental Breakfast & 12 & 10 & 0.23 & 0.20 & 0.36 \\
\hspace{2em} Greek Salad & 10 & 4 & 0.25 & 0.18 & 0.77 \\
\hspace{2em} Pasta Salad & 19 & 8 & 0.25 & 0.19 & 0.86 \\
\hspace{2em} Hot Box Pizza & 6 & 8 & 0.31 & 0.13 & 0.62 \\
\hspace{2em} Turkey Sandwich & 13 & 6 & 0.21 & 0.01 & 0.52 \\ \hline \bottomrule
\end{tabular}
}
\end{table}

\textit{Missing Key-steps (M):} This metric quantifies the proportion of omitted key-steps in each video. 
\begin{equation}
    M = 1 - \frac {\sum_{n=1}^{N} u_{n}}{KN};  \tag{A17}
\end{equation}
Values range from 0 to 1, with higher values indicating more missing steps. This measure helps assess task feasibility when certain steps are skipped.

\textit{Repeated Key-steps:} This metric captures the frequency of key-step repetition across videos:
\begin{equation}
    R = 1 - \frac {\sum_{n=1}^N u_{n}}{\sum_{n=1}^N g_{n}} \tag{A18}
\end{equation}

\vspace{-0.5em}
\subsection{Third-Person Video Perspective}

\begin{wraptable}{r}{0.5\columnwidth}
\renewcommand\thetable{A4}
\vspace{-3em}
\centering
\caption{Comparison of third-person perspectives from CMU-MMAC \cite{de2009guide_cmummac} against egocentric recordings. Egocentric view demonstrates superior alignment quality, underscoring the strength of OT in capturing first-person task dynamics.}
\vspace{-0.2em}
\label{tab:CMU-3rdperson}
\resizebox{0.43\columnwidth}{!}{%
\begin{tabular}{ccccc}
\toprule
View & \textbf{P} & \textbf{R} & \textbf{F1} & \textbf{IoU} \\ \midrule
TP (Top) & 45.5 & 49.3 & 47.3 & 32.2 \\
TP (Back) & 49.7 & 52.8 & 51.2 & 35.0 \\
TP (LHS) & 52.2 & 55.4 & 53.7 & 37.9 \\
TP (RHS) & 47.0 & 51.2 & 49.0 & 33.4 \\ \hline
\rowcolor{cyan!15} Egocentric & \textbf{61.2} & \textbf{58.4} & \textbf{59.7} & \textbf{43.7} \\ \hline \bottomrule
\end{tabular}%
}
\end{wraptable}

We evaluate REMAP on multiple third-person camera views from CMU-MMAC \cite{de2009guide_cmummac} and compare them with the egocentric setting in Table~\ref{tab:CMU-3rdperson}. The egocentric view achieves the highest F1 and IoU, likely because first-person recordings more directly capture task-relevant hand-object interactions. REMAP performs consistently across exocentric views, indicating that the learned alignment is applicable to both egocentric and exocentric procedure videos, while benefiting from views that emphasize procedural interactions.

\subsection{Quantitative results of REMAP on different subtasks across the datasets} \label{sec:app_datasets}
We report results on individual subtasks from egocentric datasets, including CMU-MMAC \cite{de2009guide_cmummac} and EGTEA-GAZE+ \cite{li2018eye_egteagaze}, in Table~\ref{tab:abs4}, and on third-person exocentric datasets such as ProceL \cite{elhamifar2020self_procel} and CrossTask \cite{zhukov2019crosstask} in Table~\ref{tab:abs5}. This evaluation spans a variety of perspectives and task domains, offering a comprehensive view of model behavior across settings. Across most subtasks, R-FPGWOT improves over R-FGWOT, showing that partial transport provides consistent gains beyond fused structural alignment alone. The improvements are especially visible in tasks with background clutter, repeated actions, or missing steps, where forcing all frames to match can introduce spurious correspondences.

\begin{table}[!ht]
\setlength\extrarowheight{2pt}
\setlength\tabcolsep{4pt} 
\renewcommand\thetable{A5}
\centering
\caption{Results on individual subtasks of egocentric datasets.}
\label{tab:abs4}
\begin{subtable}[t]{\textwidth}
\caption{EGTEA-GAZE+ \cite{li2018eye_egteagaze}}
\resizebox{\columnwidth}{!}{%
\begin{tabular}{lcccccccccccccccccccc}
\toprule
Method & \multicolumn{2}{c}{Bacon Eggs} & & \multicolumn{2}{c}{Cheeseburger} & & \multicolumn{2}{c}{Breakfast} & & \multicolumn{2}{c}{Greek Salad} & & \multicolumn{2}{c}{Pasta Salad} & & \multicolumn{2}{c}{Pizza} & & \multicolumn{2}{c}{Turkey} \\ \cline{2-3} \cline{5-6} \cline{8-9} \cline{11-12} \cline{14-15} \cline{17-18} \cline{20-21}
 & F1 & IoU & & F1 & IoU & & F1 & IoU & & F1 & IoU & & F1 & IoU & & F1 & IoU & & F1 & IoU \\ \midrule
R-FGWOT & 62.15 & 47.88 & & 63.02 & 47.37 & & 56.50 & 41.09 & & 66.78 & 51.65 & & 68.90 & 54.34 & & 53.87 & 37.61 & & 65.84 & 50.33 \\
\rowcolor{cyan!15} R-FPGWOT & 66.74 & 53.62 & & 66.98 & 51.95 & & 57.95 & 42.50 & & 66.79 & 51.65 & & 70.99 & 56.78 & & 53.97 & 37.67 & & 66.23 & 50.68 \\ \hline \bottomrule
\end{tabular}%
}
\end{subtable}

\vspace{1em}
\begin{subtable}[t]{\textwidth}
\caption{CMU-MMAC \cite{de2009guide_cmummac}}
\centering
\resizebox{0.78\columnwidth}{!}{%
\begin{tabular}{lcccccccccccccc}
\toprule
Method & \multicolumn{2}{c}{Brownie} & & \multicolumn{2}{c}{Eggs} & & \multicolumn{2}{c}{Pizza} & & \multicolumn{2}{c}{Salad} & & \multicolumn{2}{c}{Sandwich} \\ \cline{2-3} \cline{5-6} \cline{8-9} \cline{11-12} \cline{14-15}
 & F1 & IoU & & F1 & IoU & & F1 & IoU & & F1 & IoU & & F1 & IoU \\ \midrule
R-FGWOT & 58.29 & 41.67 & & 56.23 & 40.71 & & 47.27 & 31.51 & & 63.95 & 48.53 & & 65.67 & 50.37 \\
\rowcolor{cyan!15}R-FPGWOT & 58.52 & 41.92 & & 56.72 & 41.11 & & 48.00 & 32.22 & & 69.23 & 52.27 & & 66.16 & 50.89 \\ \hline \bottomrule
\end{tabular}%
}
\end{subtable}
\vspace{-1em}
\end{table}

\begin{table}[!ht]
\setlength\extrarowheight{2pt}
\setlength\tabcolsep{4pt} 
\renewcommand\thetable{A6}
\centering
\caption{Results on individual subtasks of Third-person exocentric datasets.}
\label{tab:abs5}
\begin{subtable}[t]{\textwidth}
\caption{ProceL \cite{elhamifar2020self_procel}}
\label{tab:abs5a}
\centering
\resizebox{\columnwidth}{!}{%
\begin{tabular}{>{\rule[-0.2cm]{0pt}{0.6cm}}lcccccccccccccccccc}
\toprule
\multirow{2}{*}{Methods} & \multicolumn{2}{c}{Clarinet} &  & \multicolumn{2}{c}{PB\&J Sandwich} &  & \multicolumn{2}{c}{Salmon} &  & \multicolumn{2}{c}{Jump Car} &  & \multicolumn{2}{c}{Toilet} &  & \multicolumn{2}{c}{Tire Change} \\ \cline{2-3} \cline{5-6} \cline{8-9} \cline{11-12} \cline{14-15} \cline{17-18} 
 & F1 & IoU &  & F1 & IoU &  & F1 & IoU &  & F1 & IoU &  & F1 & IoU &  & F1 & IoU \\ \midrule
R-FGWOT & 67.85 & 54.12 & & 55.69 & 40.42 & & 57.03 & 41.59 & & 67.30 & 54.85 & & 53.18 & 38.16 & & 50.85 & 35.09 \\
\rowcolor{cyan!15} R-FPGWOT & 68.48 & 54.82 & & 56.46 & 40.97 & & 58.57 & 43.25 & & 67.99 & 55.65 & & 55.27 & 40.10 & & 51.69 & 36.13 \\ \hline \bottomrule
\multirow{2}{*}{Methods} & \multicolumn{2}{c}{Tie-Tie} &  & \multicolumn{2}{c}{Coffee} &  & \multicolumn{2}{c}{iPhone Battery} &  & \multicolumn{2}{c}{Repot Plant} &  & \multicolumn{2}{c}{Chromecast} &  & \multicolumn{2}{c}{CPR} \\ \cline{2-3} \cline{5-6} \cline{8-9} \cline{11-12} \cline{14-15} \cline{17-18} 
& F1 & IoU &  & F1 & IoU &  & F1 & IoU &  & F1 & IoU &  & F1 & IoU &  & F1 & IoU \\ \midrule
R-FGWOT & 55.79 & 40.37 &  & 63.10 & 48.91 &  & 49.47 & 33.90 &  & 60.34 & 45.40 &  & 48.74 & 32.64 &  & 50.87 & 35.64 \\
\rowcolor{cyan!15} R-FPGWOT & 55.97 & 40.53 &  & 64.52 & 50.15 &  & 49.49 & 33.90 &  & 60.74 & 45.92 &  & 49.71 & 33.41 &  & 52.30 & 36.82 \\ \hline \bottomrule
\end{tabular}%
}
\end{subtable}

\vspace{1em}
\begin{subtable}[t]{\textwidth}
\caption{CrossTask \cite{zhukov2019crosstask}}
\label{tab:abs5b}
\centering
\resizebox{0.8\columnwidth}{!}{%
\begin{tabular}{>{\rule[-0.2cm]{0pt}{0.6cm}}lcccccccccccccccccc}
\toprule
\multirow{2}{*}{Methods} & \multicolumn{2}{c}{16815} &  & \multicolumn{2}{c}{23521} &  & \multicolumn{2}{c}{40567} &  & \multicolumn{2}{c}{44047} &  & \multicolumn{2}{c}{44789} &  & \multicolumn{2}{c}{53193}  \\ \cline{2-3} \cline{5-6} \cline{8-9} \cline{11-12} \cline{14-15} \cline{17-18} 
& F1 & IoU &  & F1 & IoU &  & F1 & IoU &  & F1 & IoU &  & F1 & IoU &  & F1 & IoU \\ \midrule
R-FGWOT & 64.4 & 50.0 &  & 61.3 & 46.1 &  & 58.9 & 43.7 &  & 56.7 & 41.9 &  & 60.6 & 46.6 & &  66.0 & 51.9 \\
\rowcolor{cyan!15} R-FPGWOT & 65.1 & 50.4 &  & 61.5 & 46.3 &  & 59.6 & 44.5 &  & 57.7 & 42.7 &  & 61.9 & 48.1 &  & 66.5 & 52.3 \\
\hline \bottomrule
\multirow{2}{*}{Methods} & \multicolumn{2}{c}{59684} &  & \multicolumn{2}{c}{71781} &  & \multicolumn{2}{c}{76400} &  & \multicolumn{2}{c}{77721} &  & \multicolumn{2}{c}{87706} &  & \multicolumn{2}{c}{91515}  \\ \cline{2-3} \cline{5-6} \cline{8-9} \cline{11-12} \cline{14-15} \cline{17-18} 
& F1 & IoU &  & F1 & IoU &  & F1 & IoU &  & F1 & IoU &  & F1 & IoU &  & F1 & IoU \\ \midrule
R-FGWOT & 55.0 & 40.0 &  & 62.6 & 49.5 &  & 63.0 & 48.4 &  & 64.4 & 49.9 &  & 55.3 & 39.7 &  & 58.5 & 43.2 \\
\rowcolor{cyan!15} R-FPGWOT & 56.1 & 40.1 &  & 63.6 & 50.3 &  & 63.5 & 48.9 &  & 65.2 & 50.6 &  & 55.9 & 40.3 &  & 58.9 & 43.8 \\ 
\hline \bottomrule 
\multirow{2}{*}{Methods} & \multicolumn{2}{c}{94276} &  & \multicolumn{2}{c}{95603} &  & \multicolumn{2}{c}{105222} &  & \multicolumn{2}{c}{105253} &  & \multicolumn{2}{c}{109972} &  & \multicolumn{2}{c}{113766}  \\ \cline{2-3} \cline{5-6} \cline{8-9} \cline{11-12} \cline{14-15} \cline{17-18} 
& F1 & IoU &  & F1 & IoU &  & F1 & IoU &  & F1 & IoU &  & F1 & IoU &  & F1 & IoU \\ \midrule
R-FGWOT & 57.2 & 41.7 &  & 58.5 & 43.2 &  & 60.6 & 45.2 &  & 61.3 & 46.6 &  & 62.9 & 48.5 &  & 64.2 & 49.6 \\
\rowcolor{cyan!15} R-FPGWOT & 57.6 & 42.0 &  & 58.6 & 43.3 &  & 61.6 & 46.0 &  & 62.5 & 47.8 &  & 63.9 & 49.7 &  & 65.1 & 50.5 \\ \hline \bottomrule
\end{tabular}%
}
\end{subtable}
\vspace{-1em}
\end{table}


\subsection{Additional Ablation Studies}
\label{app8}
\subsubsection{Key-Step Localization and Ordering using Graphcut Segmentation} \label{app8.1}
After learning frame embeddings using the R-FPGWOT alignment framework, we recover procedural structure by localizing key steps and inferring their temporal order. Following prior work, we cast key-step localization as a multi-label graph-cut segmentation problem \cite{greig1989exact}, where the node set includes $K$ terminal nodes representing key steps and non-terminal nodes corresponding to frame embeddings. Given frame embeddings $\{\boldsymbol z_i\}_{i=1}^T$ and key-step prototypes $\{\boldsymbol c_k\}_{k=1}^K$, we construct a graph $G=(\mathcal V,\mathcal E)$ with node set $\mathcal V=\{1,\ldots,T\}$ and label space $\mathcal L=\{1,\ldots,K\}$. The segmentation $\boldsymbol y\in\mathcal L^T$ is obtained by minimizing the Potts-model energy:
\vspace{-0.6em}
\begin{equation}
    E(\boldsymbol y)
    \;=\;
    \underbrace{\sum_{i=1}^T D_i(y_i)}_{\text{T-links: data term}}
    \;+\;
    \underbrace{\beta \sum_{(i,j)\in\mathcal{E}_t} w_{ij}\,\mathbb{I}[y_i \neq y_j]}_{\text{N-links: temporal smoothness}}. \tag{A19}
    \label{eqn:graphcut}
\end{equation}

\vspace{-1em}
where $w_{ij}$ penalizes label changes between temporally adjacent frames.

Here, the T-links implement the data term:
\begin{equation}
    D_i(k) \;=\; \|\boldsymbol z_i - \boldsymbol c_k\|_2^2, \tag{A20}
\end{equation}
which encourages frame $i$ to attach to the key-step prototype with the most similar embedding, enforcing structural consistency between clusters and the learned embedding geometry. 

\noindent The N-links connect temporally adjacent frames $(i,j) \in \mathcal{E}_t$ (typically $j=i+1$) with weights:
\begin{equation}
    w_{ij} \;=\; \exp\!\Big(-\frac{\|\boldsymbol z_i - \boldsymbol z_j\|_2^2} {2\sigma^2}\Big), \tag{A21}
\end{equation}
so that label changes between nearby frames with similar embeddings incur a higher penalty. This formulation promotes temporally contiguous segments while allowing boundaries at points of significant embedding change. The resulting submodular energy is approximately minimized using $\alpha$-Expansion \cite{boykov2002fast}, producing piecewise-constant key-step segments consistent with both embedding similarity and temporal continuity. The same graph-cut decoder is used only after embedding learning and does not affect the OT objective or training gradients.

To infer the step order, we leverage Hungarian-algorithm based alignment matches \cite{chowdhury2024opel}. Frame timestamps are first normalized within each video, and the mean normalized time of the frames assigned to each cluster is computed. The sorting of clusters by these means yields a predicted step sequence for each video. Finally, we aggregate sequences across all videos of the same task, rank them by frequency of occurrence, and select the most frequent ordering as the canonical procedure. This post-hoc pipeline reliably recovers both salient steps and their temporal organization without influencing the underlying OT alignment.

\begin{algorithm}[H]
\caption{Temporal Ordering of Key Steps}
\label{alg:temporal_order}
\begin{algorithmic}[1]
\Require $R$: predicted key-step assignment for each frame, $k$: number of key steps
\Ensure $\text{indices}$: sequential order of tasks
\State $M \gets \text{len}(R)$ \Comment{Number of frames}
\State $T \gets \frac{\{1,2,\dots,M\}}{M}$ \Comment{Normalized timestamps}
\State Initialize $cluster\_time \gets \mathbf{0}_k$
\For{$i=1$ \textbf{to} $k$}
    \State $cluster\_time[i] \gets \text{mean}(T[R==i])$
\EndFor
\State $\_, indices \gets \text{sort}(cluster\_time)$
\State \Return $indices$
\end{algorithmic}
\vspace{0.3em}
\noindent\textbf{Example:} \\
\texttt{\textbf{Sample Input (R):}} 
\texttt{[6, 2, 1, 3, 5, 1, 1, 0, 0, 6, 4, 4, 6, 1, 2, 3, 0, 4, 0, 4, 5, 5, 3, 1, 3, 2, 0, 4, 3, 6, 0, 1, 2, 4, 2, 3, 5, 4, 6, 2, 5, 1, 2, 4, 3, 2, 2, 3]} \\[0.3em]
\texttt{\textbf{Sample Output (indices):}} 
\texttt{[1, 0, 6, 5, 4, 3, 2]}
\end{algorithm}

\subsubsection{Choice of Key-Step $K$} \label{app8.2}

\begin{wrapfigure}{r}{0.5\columnwidth}
    \renewcommand{\thefigure}{A1}
    \centering
    \vspace{-2em}
    \includegraphics[width=0.48\columnwidth]{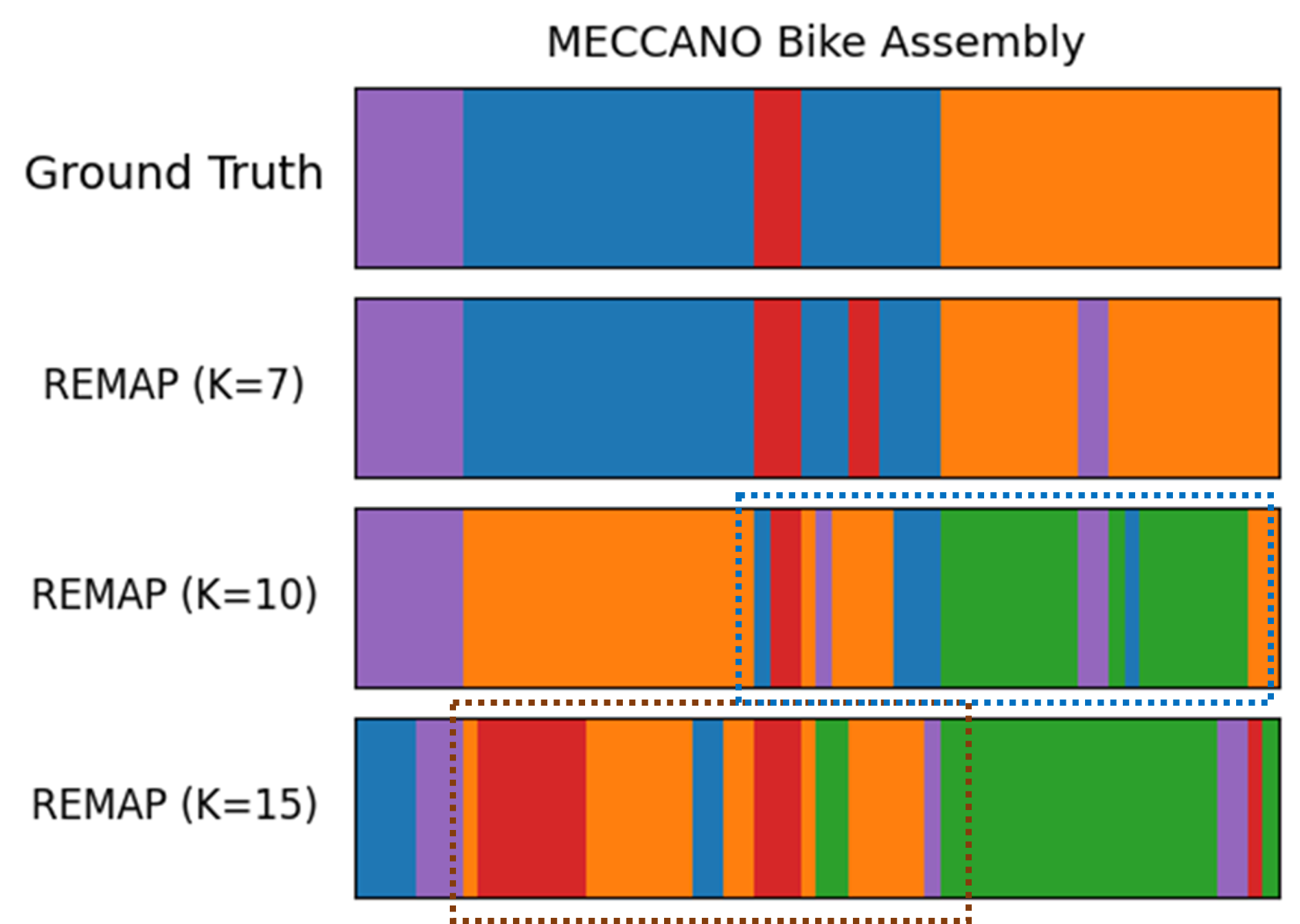}
    \caption{Ablation study on the choice of $K$. 
    With $K=7$, the model achieves the best balance between capturing essential task boundaries and avoiding over-segmentation. Increasing $K$ leads to more fragmented and jittery segmentations.
    }
    \label{fig:ablation_k}
\end{wrapfigure}

REMAP uses a fixed global value $K=7$ across all datasets to maintain a fully unsupervised and dataset-agnostic setting. This choice approximately matches the average number of semantically distinct procedural stages observed across the evaluated datasets and provides a stable balance between under-segmentation and excessive fragmentation. Importantly, the discovered clusters represent latent procedural stages rather than strict one-to-one reconstructions of all atomic annotations. In datasets with highly fine-grained labels, such as MECCANO, multiple related annotations may naturally collapse into a shared higher-level procedural cluster while still preserving the dominant procedural structure. We further evaluate the effect of varying $K$ on segmentation quality. Small values of $K$ under-segment the sequence by merging distinct task boundaries, whereas larger values (e.g., $K=10$ or $15$) over-segment the procedure into short and unstable fragments, introducing temporal jitter and reducing interpretability (Fig.~\ref{fig:ablation_k}). Since procedurally similar actions (e.g., pouring oil vs. pouring water) may reasonably belong to the same latent stage, increasing $K$ beyond the underlying semantic granularity produces near-duplicate clusters and degrades segmentation quality. Overall, $K=7$ provides the most robust and semantically coherent decomposition across datasets.

\subsubsection{Prior Distributions and Sensitivity to the Laplace Scale Parameter}
\label{sec:priors_and_laplace_scale}
\begin{figure}[!ht]
\renewcommand{\thefigure}{A2}
  \centering
    \vspace{-1em}
   \includegraphics[width=0.8\linewidth]{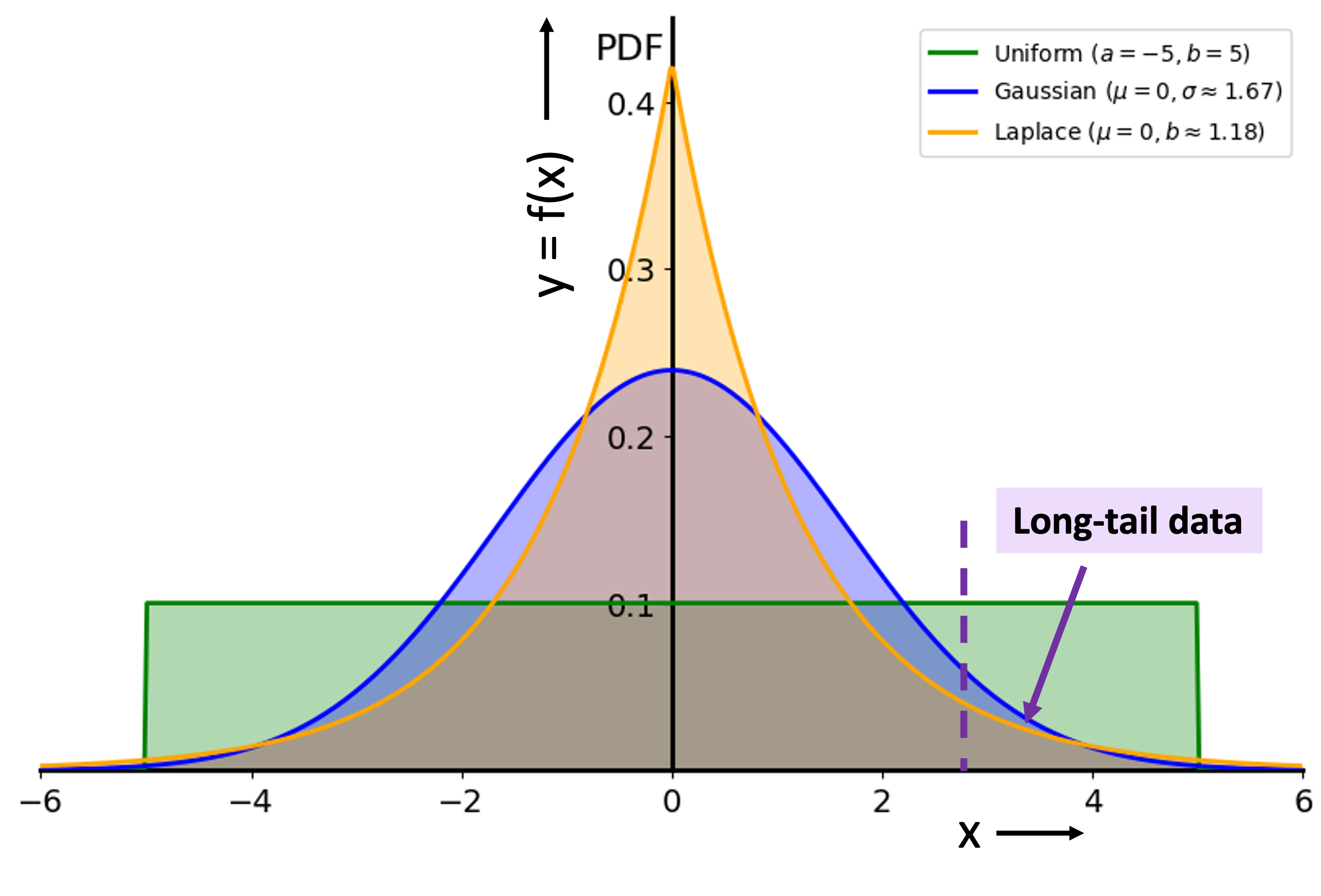}
   \vspace{-1em}
   \caption{Importance of choosing Laplace distribution as a prior over Gaussian and Uniform distribution.}
  \label{fig:priors}
\end{figure}

In R-FPGWOT, temporal and optimality structure is encoded through a parametric prior over pairwise frame distances, implemented as a symmetric kernel. We evaluate three prior choices -- Uniform, Gaussian, and Laplace (Fig.~\ref{fig:priors}) -- which all impose temporal locality but differ in how strongly they penalize deviations from diagonal alignment. The Uniform prior (Eq.~\ref{eq:a20}) enforces a hard locality window by assigning equal mass within a fixed range and zero elsewhere, but lacks discrimination among alignments inside the window. The Gaussian prior (Eq.~\ref{eq:a21}) produces a sharply peaked kernel with rapidly decaying tails, favoring near-diagonal matches while strongly penalizing moderate temporal shifts and non-monotonic correspondences. In contrast, the Laplace prior (Eq.~\ref{eq:a22}) combines a sharp central mode with heavier tails, maintaining a strong diagonal bias while allowing non-negligible mass for moderately misaligned frames. This property makes it particularly well-suited to realistic egocentric videos, where small temporal jitter and occasional non-monotonic transitions are common.

\vspace{-1em}
\begin{equation}
    \boldsymbol{Q}(i, j) = f(x; a, b) = \begin{cases}
    \frac{1}{b-a} & \text{if } a \leq x \leq b, \\
    0 & \text{otherwise}.
    \end{cases}
\tag{A22} \label{eq:a20}
\end{equation}

\begin{equation}
    \boldsymbol{Q}(i, j) = \mathcal{N}(x; \mu, \sigma^2) = \frac{1}{\sqrt{2 \pi \sigma^2}} \exp\left( -\frac{(x - \mu)^2}{2 \sigma^2} \right)
\tag{A23} \label{eq:a21}
\end{equation}

\begin{equation}
    \boldsymbol{Q}(i, j) = f(x; \mu, b) = 
    \begin{cases}
        \dfrac{1}{2b} \exp\!\left(-\dfrac{|x - \mu|}{b}\right) & \text{if } -\infty < x < \infty, \\
        0 & \text{otherwise}.
    \end{cases}
\tag{A24} \label{eq:a22}
\end{equation}

Table~\ref{tab:prior_dist} reports F1/IoU performance across EgoProceL datasets for all three priors. The Laplace prior, used by default in R-FPGWOT, consistently achieves the best or tied-best results, outperforming both Uniform and Gaussian variants. 
Together with the convergence behavior discussed in the main text, these results motivate our choice of the Laplace prior, as it provides a favorable balance between diagonal concentration and robustness to temporal variability.

\vspace{-0.5em}
\begin{table}[!ht]
\renewcommand\thetable{A7}
\centering
\caption{Ablation on the choice of prior distribution (Uniform, Gaussian, Laplace) for the temporal and optimality kernels in R-FPGWOT on EgoProceL. We report F1/IoU (\%) for each dataset.}
\vspace{0.4em}
\label{tab:prior_dist}
\resizebox{0.9\columnwidth}{!}{%
\begin{tabular}{lccccccccccc}
\toprule
 & \multicolumn{11}{c}{EgoProceL} \\ \cline{2-12} 
 & \multicolumn{2}{c}{CMU-MMAC} &  & \multicolumn{2}{c}{EGTEA-GAZE+} &  & \multicolumn{2}{c}{MECCANO} &  & \multicolumn{2}{c}{EPIC-Tents} \\ 
\cline{2-3} \cline{5-6} \cline{8-9} \cline{11-12}
 & F1 & IoU &  & F1 & IoU &  & F1 & IoU &  & F1 & IoU \\ \midrule
Uniform& 53.0 & 37.0 & & 57.5 & 42.5 & & 53.0 & 36.0 & & 33.0 & 18.5 \\
Gaussian& 57.0 & 41.0 & & 61.5 & 46.5 & & 57.0 & 40.0 & & 37.0 & 22.5 \\
\rowcolor{cyan!15} Laplace & \textbf{59.7} & \textbf{43.7} & & \textbf{64.2} & \textbf{49.3} & & \textbf{59.6} & \textbf{42.7} & & \textbf{39.8} & \textbf{25.0} \\ \hline \bottomrule
\end{tabular}
}
\vspace{-1em}
\end{table}

\begin{figure}[!ht]
    \centering
    \begin{minipage}{0.54\linewidth}
        \centering
        \renewcommand\thetable{A8}
        \captionof{table}{Ablation on the Laplace scale parameter \(b\) for the temporal and optimality priors. We report F1/IoU (\%) on CMU-MMAC, EGTEA-GAZE+, and MECCANO.}
        \label{tab:laplace_scale}
        \resizebox{\columnwidth}{!}{%
        \begin{tabular}{lcccccc}
            \toprule
            \multirow{2}{*}{$b_{\text{Laplace}}$} & \multicolumn{2}{c}{CMU-MMAC} & \multicolumn{2}{c}{EGTEA-GAZE+} & \multicolumn{2}{c}{MECCANO} \\
            & F1 & IoU & F1 & IoU & F1 & IoU \\
            \midrule
            1.0 & 57.64 & 42.18 & 63.04 & 47.95 & 57.71 & 40.94 \\
            1.5 & 58.23 & 42.98 & 63.23 & 48.57 & 57.96 & 41.36 \\
            \cellcolor{cyan!15} 2.0 & \cellcolor{cyan!15}\textbf{59.73} & \cellcolor{cyan!15}\textbf{43.68} & \cellcolor{cyan!15}\textbf{64.23} & \cellcolor{cyan!15}\textbf{49.26} & 58.52 & 41.57  \\
            2.5 & 58.10 & 42.72 & 62.10 & 48.31 & 58.61 & 42.18 \\
            \cellcolor{red!15} 3.0 & 57.98 & 42.12 & 61.98 & 47.75 & \cellcolor{red!15}\textbf{59.61} & \cellcolor{red!15} \textbf{42.67}  \\
            3.5 & 57.83 & 41.95 & 61.84 & 47.23 & 58.28 & 41.71 \\
            \hline \bottomrule
        \end{tabular}
        }
    \end{minipage}%
    \hfill%
    \renewcommand{\thefigure}{A3}
    \begin{minipage}{0.44\linewidth}
        \centering
        \includegraphics[width=1.05\linewidth]{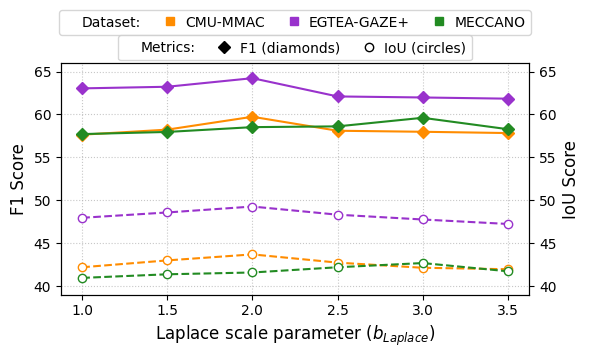}
        \vspace{-1em}
        \caption{Sensitivity of R-FPGWOT to the Laplace scale parameter \(b\).}
        \label{fig:laplace_scale}
    \end{minipage}
\end{figure}

\paragraph{Sensitivity to the Laplace scale parameter \(b\).}
The Laplace scale $b$ controls how quickly the temporal prior decays away from the diagonal. Intuitively, small \(b\) produces an overly peaked kernel that restricts alignments to a narrow local neighborhood, while very large \(b\) flattens the kernel and weakens the temporal guidance, approaching a weakly informative prior. We evaluate \(b \in \{1, 1.5, 2, 2.5, 3, 3.5\}\) and report F1/IoU on CMU-MMAC, EGTEA-GAZE+, and MECCANO in Table~\ref{tab:laplace_scale}. Fig.~\ref{fig:laplace_scale} show that performance is stable across a broad range of $b$, with the best values around $b=2$ for CMU-MMAC and EGTEA-GAZE+ and $b=3$ for MECCANO. This supports using a small dataset-level choice of $b$ rather than extensive tuning.

\begin{figure}[!ht]
    \renewcommand{\thefigure}{A4}
    \centering
    \includegraphics[width=0.82\linewidth]{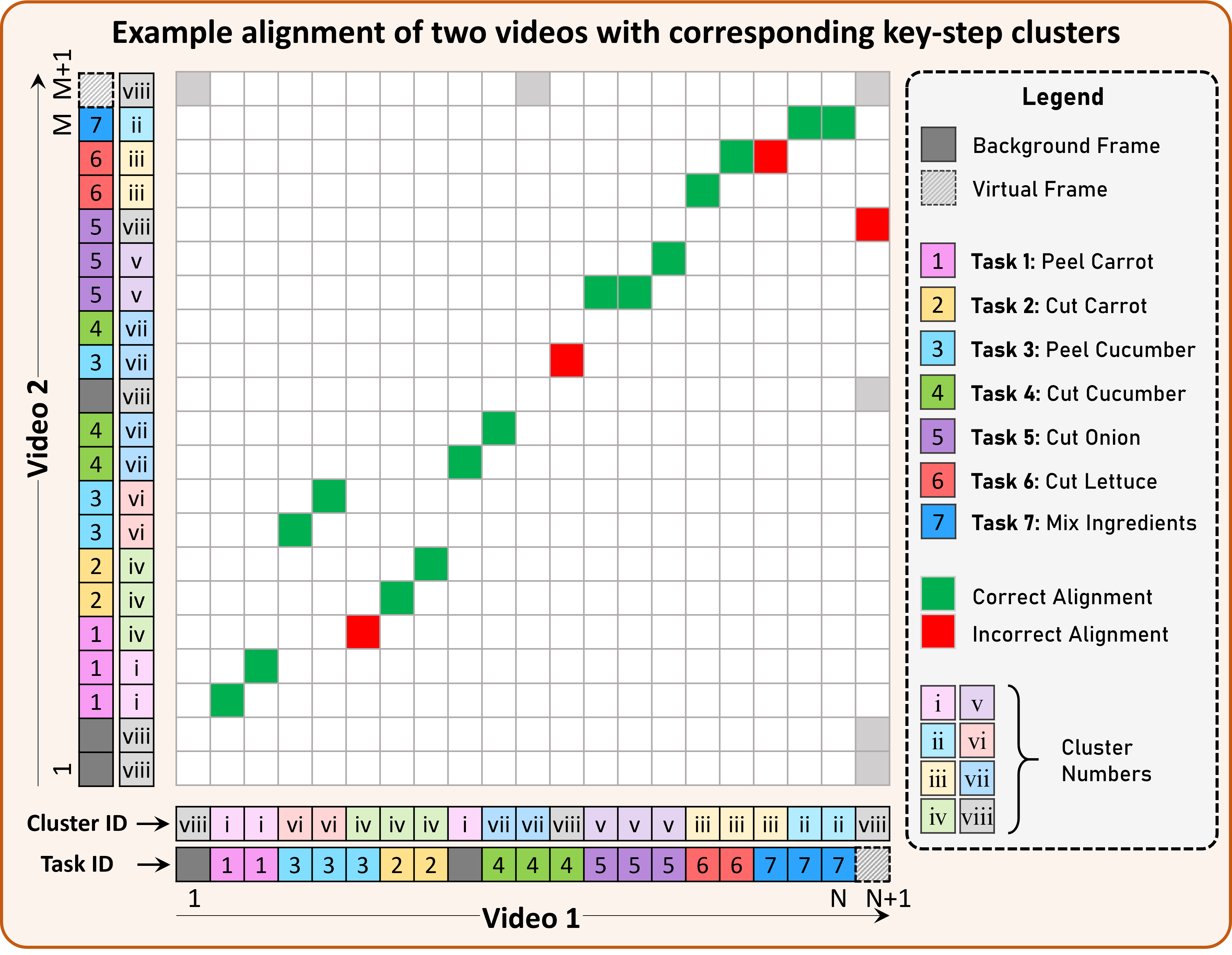}
    \caption{Illustration of sequence alignment of two `salad making' videos with different temporal dynamics using our framework `REMAP'. Despite variations in execution speed, corresponding action frames are matched accurately, thereby managing redundancy and robustness of our method.}
    \label{fig:alignment_example}
    \vspace{-1.5em}
\end{figure}

\subsubsection{Sequence Alignment Robustness} \label{app8.4}
To further assess robustness, we evaluate alignment between video pairs with different temporal variations. As shown in Fig.~\ref{fig:alignment_example}, REMAP aligns corresponding action frames despite temporal stretching and speed variation. Redundant or prolonged regions are handled without disrupting key-step correspondences, confirming that the learned transport plan preserves coherent procedural alignment across temporally diverse demonstrations.

\subsubsection{Additional Applications} 
\label{app7}
\vspace{-0.5em}
Because REMAP learns frame-level correspondences across videos of the same task, it can support downstream applications such as procedure monitoring, assistive guidance, robotic imitation, annotation transfer, and anomaly detection. In procedure monitoring, the system can automatically verify whether each key step is performed correctly, flagging errors or deviations. For assistive guidance, it can localize the current step in real time and suggest the next, serving as an intelligent instruction system. In robotic automation, the framework learns procedural knowledge directly from observation, allowing robots to replicate tasks without explicit programming.

\begin{figure}[!ht]
\renewcommand{\thefigure}{A5}
\centering
\includegraphics[width=0.82\linewidth]{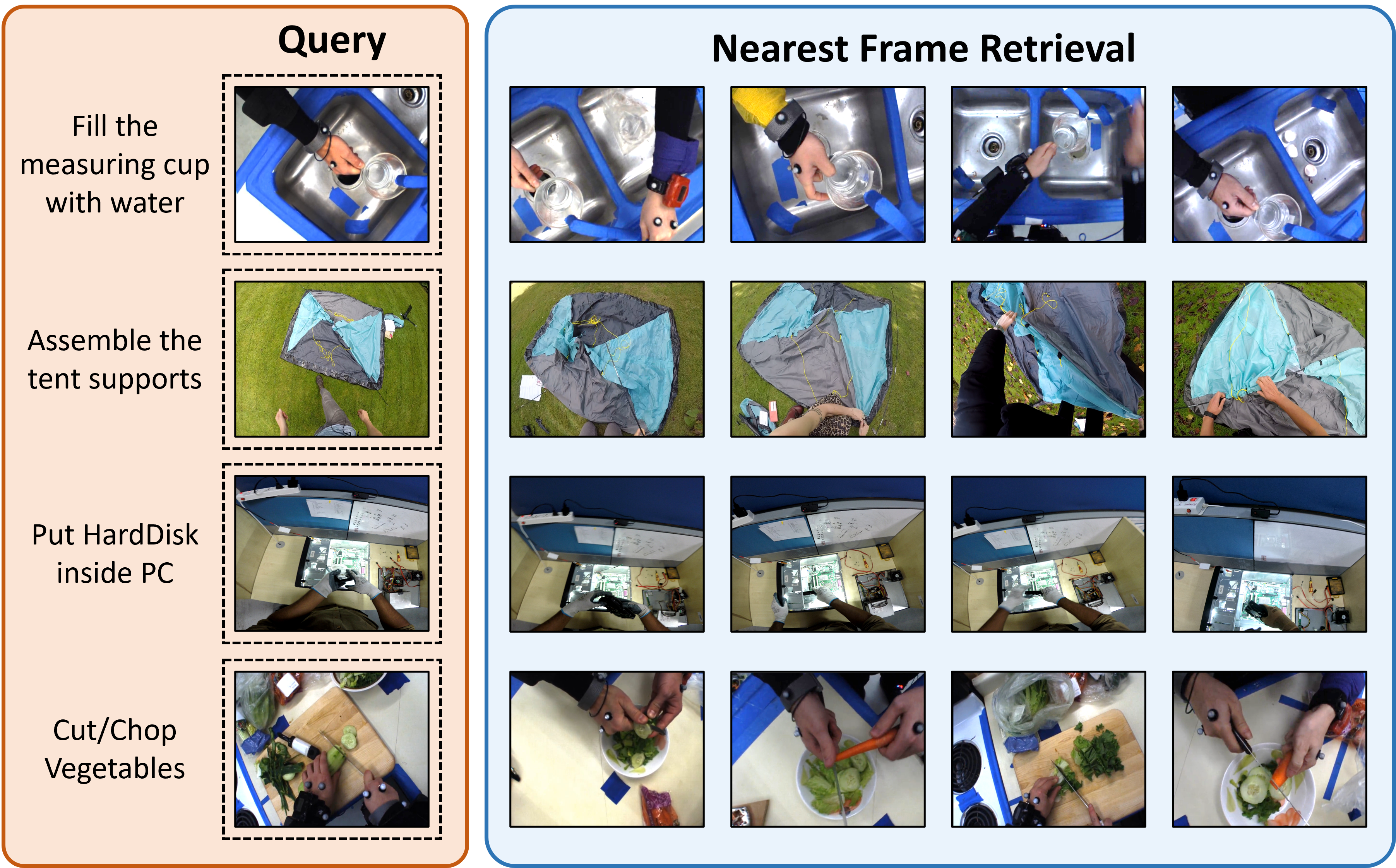}
\caption{Nearest-neighbor retrieval in the embedding space enables precise frame-level alignment across tasks.}
\label{fig:ap1}
\end{figure}

Beyond execution, the model also supports cross-modal transfer: annotations or cues (e.g., text or audio) can be propagated across aligned videos. The embedding space further enables fine-grained retrieval and anomaly detection. Nearest-neighbor search surfaces frames corresponding to specific actions, while deviations from expected trajectories indicate abnormal behavior, ensuring correct procedural order. Figure~\ref{fig:ap1} illustrates these capabilities: retrieving filled-container frames in water-filling (Row 1), distinguishing pre- vs. post-assembly in tent assembly (Row 2), identifying hard disk insertion in PC assembly (Row 3), and finding chopping actions across different vegetables (Row 4). These results suggest that the learned alignment space may be useful beyond key-step localization.

\subsubsection{Sensitivity Analyses of Other Hyperparameters} \label{app8.5}
\begin{figure}[!ht]
\renewcommand{\thefigure}{A6}
  \centering
   \includegraphics[width=0.98\linewidth]{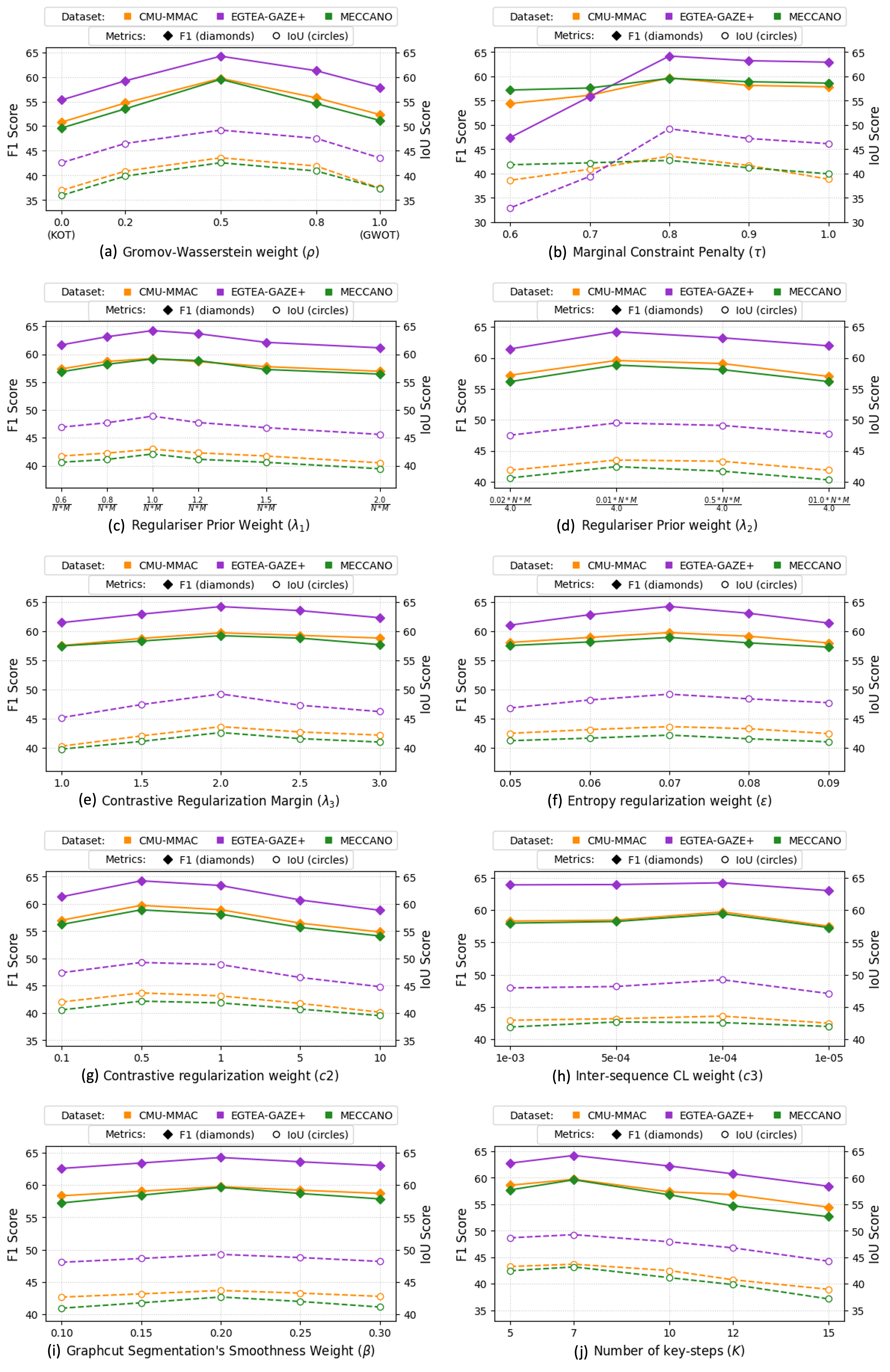}
   \caption{Sensitivity analysis of the various hyperparameters used in \textit{REMAP}}.
  \label{fig:hyper}
\end{figure}

We further analyze the sensitivity of \textit{REMAP} to several key hyperparameters that control distinct and interpretable aspects of the alignment. Fig.~\ref{fig:hyper} reports sensitivity analyses for the main REMAP hyperparameters on CMU-MMAC, EGTEA-GAZE+, and MECCANO. The partial transport strength $\tau$ determines how much unmatched or background content is allowed to remain unaligned; smaller values allow more mass to be absorbed by the virtual sink, while larger values recover the balanced OT setting. The temperatures $\lambda_2$ and $\lambda_3$ control the sharpness of the Gibbs kernel and primarily affect convergence speed rather than the alignment structure. The structural regularization weight $\lambda_1$ and Laplace priors influence the smoothness and near-diagonal bias of the transport plan, but do not enforce strict monotonicity. Other hyperparameters include the Gromov-Wasserstein weight $\rho$, contrastive weights $c_2$ and $c_3$, entropy regularization $\epsilon$, graph-cut smoothness $\beta$, and the number of key steps $K$.

Across all settings, REMAP exhibits stable behavior across broad ranges, supporting the use of a fixed configuration across datasets. Performance curves are generally smooth and unimodal, with optima near $\rho \approx 0.5$ (balancing KOT and GWOT), $\tau \approx 0.8$, $\lambda_1 \approx \frac{1.0}{N+M}$, $\lambda_2 \approx \frac{0.1*N*M}{4.0}$, $\lambda_3 \approx 2.0$, $c_2 \approx 0.5$, $c_3 \approx 10^{-4}$, $\epsilon \approx 0.07$, and $\beta \approx 0.2$. The most influential parameters are the partial transport strength $\tau$, the KOT--GWOT trade-off $\alpha$, and the Laplace scale $b$, which directly control unmatched mass, structural alignment, and temporal locality, respectively. We therefore fix these values as the default configuration in all experiments. Other parameters, including contrastive weights, entropy regularization, and graph-cut smoothness, have smaller effects within reasonable ranges. Performance degrades mainly under extreme settings, such as near-balanced transport or overly weak temporal structure.

\newpage
\section*{NeurIPS Paper Checklist}

\begin{enumerate}

\item {\bf Claims}
    \item[] Question: Do the main claims made in the abstract and introduction accurately reflect the paper's contributions and scope?
     \item[] Answer: \answerYes{}
    \item[] Justification: The abstract and introduction state the main claims of the paper, including the introduction of a novel approach for procedure learning, REMAP, and its contributions: partial structure-aware OT, virtual sink handling of unmatched frames, and demonstrating significant improvement through empirical results on benchmark datasets. These claims are supported by the methodology, main experiments, ablations, and appendix analyses. The limitations and future directions are also mentioned, providing a comprehensive overview of the paper's contributions and scope.

\item {\bf Limitations}
    \item[] Question: Does the paper discuss the limitations of the work performed by the authors?
    \item[] Answer: \answerYes{}
    \item[] Justification: The paper discusses limitations related to OT scalability, temporal subsampling, very long sequences, fine-grained transitions, and potential fragmentation near ambiguous boundaries. Additional runtime and subsampling analyses are provided in the appendix. These discussions demonstrate the authors' awareness of the limitations of their work and their intention to address them in future research.
    

\item {\bf Theory assumptions and proofs}
    \item[] Question: For each theoretical result, does the paper provide the full set of assumptions and a complete (and correct) proof?
     \item[] Answer: \answerYes{}
    \item[] Justification: The paper includes theoretical contributions, such as the introduction of the REMAP framework and its associated assumptions. The appendix states the assumptions for the R-FPGWOT optimization, including positive kernels, positive marginals, PSD temporal kernels, and unbalanced Sinkhorn convergence conditions. The derivation and optimization details are provided in App.~\ref{app:proof_fpgwot}. Additionally, the empirical results are supported by theoretical underpinnings regarding optimal transport theory, ensuring completeness and correctness.
    

    \item {\bf Experimental result reproducibility}
    \item[] Question: Does the paper fully disclose all the information needed to reproduce the main experimental results of the paper to the extent that it affects the main claims and/or conclusions of the paper (regardless of whether the code and data are provided or not)?
    \item[] Answer: \answerYes{}
    \item[] Justification: The paper specifies datasets, evaluation metrics, feature extraction, model architecture, training setup, hyperparameters, clustering procedure, and compute resources. Full hyperparameter settings are reported in App.~\ref{app2}, ensuring that other researchers can replicate the results. Our code is withheld until acceptance to preserve anonymity or pending publication. 

\item {\bf Open access to data and code}
    \item[] Question: Does the paper provide open access to the data and code, with sufficient instructions to faithfully reproduce the main experimental results, as described in supplemental material?
    \item[] Answer: \answerNo{}
    \item[] Justification: The paper uses publicly available benchmark datasets and cites their sources. The code is currently withheld until acceptance to preserve anonymity and will be provided upon acceptance.

\item {\bf Experimental setting/details}
    \item[] Question: Does the paper specify all the training and test details (e.g., data splits, hyperparameters, how they were chosen, type of optimizer) necessary to understand the results?
    \item[] Answer: \answerYes{}
    \item[] Justification: The experimental setup describes datasets, metrics, baselines, feature extraction, optimization, sampling strategy, and evaluation protocol. Additional hyperparameters and implementation details are provided in App.~\ref{app2}.

\item {\bf Experiment statistical significance}
    \item[] Question: Does the paper report error bars suitably and correctly defined or other appropriate information about the statistical significance of the experiments?
     \item[] Answer: \answerNo{}
    \item[] Justification: Main results are averaged over multiple runs where stated, but we do not report full error bars or confidence intervals for all experiments due to computational cost. The experimental setup is in line with the SOTA works in PL \cite{bansal2022egoprocel_pcass,bansal2024united,chowdhury2024opel,mahmood2025procedure}. Sensitivity analyses and ablations are included to assess robustness.

\item {\bf Experiments compute resources}
    \item[] Question: For each experiment, does the paper provide sufficient information on the computer resources (type of compute workers, memory, time of execution) needed to reproduce the experiments?
    \item[] Answer: \answerYes{}
    \item[] Justification: The appendix reports compute resources, including GPU type, memory usage, batch size, and approximate training time. See App.~\ref{app3}.
    
\item {\bf Code of ethics}
    \item[] Question: Does the research conducted in the paper conform, in every respect, with the NeurIPS Code of Ethics \url{https://neurips.cc/public/EthicsGuidelines}?
    \item[] Answer: \answerYes{}
    \item[] Justification: The work uses existing public research datasets and does not introduce human-subject data collection, sensitive attributes, or high-risk deployment. The research conforms to the NeurIPS Code of Ethics.

\item {\bf Broader impacts}
    \item[] Question: Does the paper discuss both potential positive societal impacts and negative societal impacts of the work performed?
    \item[] Answer: \answerYes{}
    \item[] Justification: The paper discusses potential applications such as procedure monitoring, assistive guidance, robotic imitation, annotation transfer, and anomaly detection. Potential limitations and risks from incorrect alignment or missed procedural steps are also acknowledged.
    
\item {\bf Safeguards}
    \item[] Question: Does the paper describe safeguards that have been put in place for responsible release of data or models that have a high risk for misuse (e.g., pre-trained language models, image generators, or scraped datasets)?
    \item[] Answer: \answerNA{} 
    \item[] Justification: The paper does not release high-risk models such as generative models, language models, surveillance systems, or scraped datasets. The proposed method is an alignment framework evaluated on existing procedure-learning benchmarks. However, we acknowledge that similar techniques could be misused in contexts such as unauthorized behavioral surveillance, worker monitoring, or profiling, particularly if applied without consent or transparency. Additionally, biases present in training data may lead to unequal performance across different user groups or environments. To mitigate these risks, we emphasize the importance of responsible deployment, including transparency in data collection, fairness-aware evaluation across diverse populations, and adherence to privacy-preserving practices in future.

\item {\bf Licenses for existing assets}
    \item[] Question: Are the creators or original owners of assets (e.g., code, data, models), used in the paper, properly credited and are the license and terms of use explicitly mentioned and properly respected?
    \item[] Answer: \answerYes{}
    \item[] Justification: The paper uses existing benchmark datasets and pretrained backbones, and cites the original sources for all datasets and baselines. We follow the intended research use of these assets.

\item {\bf New assets}
    \item[] Question: Are new assets introduced in the paper well documented and is the documentation provided alongside the assets?
    \item[] Answer: \answerNo{}
    \item[] Justification: The paper introduces a new method but does not release a new dataset or benchmark asset.

\item {\bf Crowdsourcing and research with human subjects}
    \item[] Question: For crowdsourcing experiments and research with human subjects, does the paper include the full text of instructions given to participants and screenshots, if applicable, as well as details about compensation (if any)? 
    \item[] Answer: \answerNA{}
    \item[] Justification: The paper does not involve new crowdsourcing experiments or new human-subject data collection. All experiments use existing public datasets.

\item {\bf Institutional review board (IRB) approvals or equivalent for research with human subjects}
    \item[] Question: Does the paper describe potential risks incurred by study participants, whether such risks were disclosed to the subjects, and whether Institutional Review Board (IRB) approvals (or an equivalent approval/review based on the requirements of your country or institution) were obtained?
    \item[] Answer: \answerNA{}
    \item[] Justification: The work does not involve new human-subject experiments or data collection by the authors. It only uses existing publicly available benchmarks.

\item {\bf Declaration of LLM usage}
    \item[] Question: Does the paper describe the usage of LLMs if it is an important, original, or non-standard component of the core methods in this research? Note that if the LLM is used only for writing, editing, or formatting purposes and does \emph{not} impact the core methodology, scientific rigor, or originality of the research, declaration is not required.
    \item[] Answer: \answerNA{}
    \item[] Justification: LLMs are not used as part of the core method, experimental pipeline, or scientific contribution.

\end{enumerate}

\end{document}